\documentclass{article} 
\usepackage{colm2024_conference}

\usepackage{microtype}
\usepackage{hyperref}
\usepackage{url}
\usepackage{booktabs}
\definecolor{darkblue}{rgb}{0, 0, 0.5}
\hypersetup{colorlinks=true, citecolor=darkblue, linkcolor=darkblue, urlcolor=darkblue}

\usepackage{enumitem}
\usepackage{graphicx}
\usepackage{url}
\usepackage{booktabs}
\usepackage{multirow}
\usepackage{tcolorbox}
\usepackage{graphicx}
\usepackage{subcaption}
\usepackage{etoolbox}
\usepackage{booktabs}
\usepackage{bbding} 
\let\classAND\AND
\let\AND\relax
\usepackage{algorithmic}

\let\AND\classAND
\AtBeginEnvironment{algorithmic}{\let\AND\algoAND}

\usepackage{algorithm}
\usepackage{algorithmic}

\usepackage{soul}
\usepackage{url}
\usepackage{booktabs}
\usepackage{arydshln}
\usepackage{nicefrac}
\usepackage{multirow}
\usepackage{tabularx}
\usepackage{amsmath}
\usepackage{amssymb}
\usepackage{amsfonts}
\usepackage{mathtools}
\usepackage{fontawesome}

\usepackage{xcolor,colortbl}
\usepackage[export]{adjustbox}

\usepackage{wrapfig}

\usepackage{tcolorbox}
\tcbuselibrary{listings,breakable}
\tcbset{listing engine=listings,colframe=black,colback=white,size=small}

\usepackage{upquote}
\definecolor{dkgreen}{rgb}{0,0.6,0}
\definecolor{gray}{rgb}{0.5,0.5,0.5}
\definecolor{mauve}{rgb}{0.58,0,0.82}
\usepackage{xspace}

\usepackage{setspace}
\usepackage{xcolor,colortbl}
\usepackage{tcolorbox}
\usepackage{subcaption}
\usepackage{graphicx}

\newcommand{\qwencoder}{Qwen2.5-Coder\xspace}
\newcommand{\fst}[1]{\textbf{#1}}

\newcommand{\huggingface}{\raisebox{-1.5pt}{\includegraphics[height=1.05em]{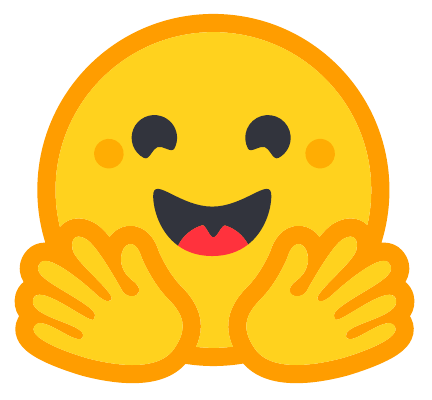}}\xspace}
\newcommand{\github}{\raisebox{-1.5pt}{\includegraphics[height=1.05em]{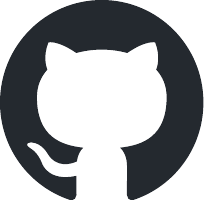}}\xspace}

\newcommand{\graybg}{\rowcolor{gray!20}}

\newcommand{\grayt}{\color{gray!80}}

\title{\centering \LARGE Qwen2.5-Coder Technical Report}

\author{
Binyuan Hui*\hspace{3mm} 
Jian Yang*\hspace{3mm} 
Zeyu Cui*\hspace{3mm} 
Jiaxi Yang*\hspace{3mm} 
\\
\normalsize{}
Dayiheng Liu\hspace{3mm} 
Lei Zhang\hspace{3mm} 
Tianyu Liu\hspace{3mm} 
Jiajun Zhang\hspace{3mm} 
Bowen Yu\hspace{3mm} 
Keming Lu\hspace{3mm}
\\
\normalsize{} 
Kai Dang\hspace{3mm} 
Yang Fan\hspace{3mm}
Yichang Zhang\hspace{3mm}
An Yang\hspace{3mm}
Rui Men\hspace{3mm}
Fei Huang\hspace{3mm}
\\ 
\normalsize{} 
Bo Zheng\hspace{3mm}
Yibo Miao\hspace{3mm}
Shanghaoran Quan\hspace{3mm} 
Yunlong Feng\hspace{3mm}
\hspace{3mm}
\\
\normalsize{} 
Xingzhang Ren\hspace{3mm}
Xuancheng Ren\hspace{3mm}
Jingren Zhou\hspace{3mm}
Junyang Lin$^{\dag}$
\\
\vspace{7mm}
\textbf{Qwen Team\hspace{3mm}Alibaba Group}
\vspace{-7mm}
}

\colmfinalcopy
\begin{document}

\maketitle

\begin{center}
\vspace{-1cm}
\begin{tabular}{rl}
\huggingface & \url{https://hf.co/Qwen/Qwen2.5-Coder-32B-Instruct}\\
\github & \url{https://github.com/QwenLM/Qwen2.5-Coder}\\
\end{tabular}
\end{center}

\begin{abstract}
{\let\thefootnote\relax\footnotetext{$^*$Equal core contribution, $^\dag$Corresponding author}}
In this report, we introduce the Qwen2.5-Coder series, a significant upgrade from its predecessor, CodeQwen1.5. This series includes six models: Qwen2.5-Coder-(0.5B/1.5B/3B/7B/14B/32B). As a code-specific model, Qwen2.5-Coder is built upon the Qwen2.5 architecture and continues pretrained on a vast corpus of over 5.5 trillion tokens. Through meticulous data cleaning, scalable synthetic data generation, and balanced data mixing, Qwen2.5-Coder demonstrates impressive code generation capabilities while retaining general and math skills.
These models have been evaluated on a wide range of code-related tasks, achieving state-of-the-art (SOTA) performance across more than 10 benchmarks, including code generation, completion, reasoning, and repair, consistently outperforming larger models of the same model size.
We believe that the release of the Qwen2.5-Coder series will advance research in code intelligence and, with its permissive licensing, support wider adoption by developers in real-world applications.
\end{abstract}

\begin{figure}[htbp]
    \centering
    \includegraphics[width=0.7\columnwidth]{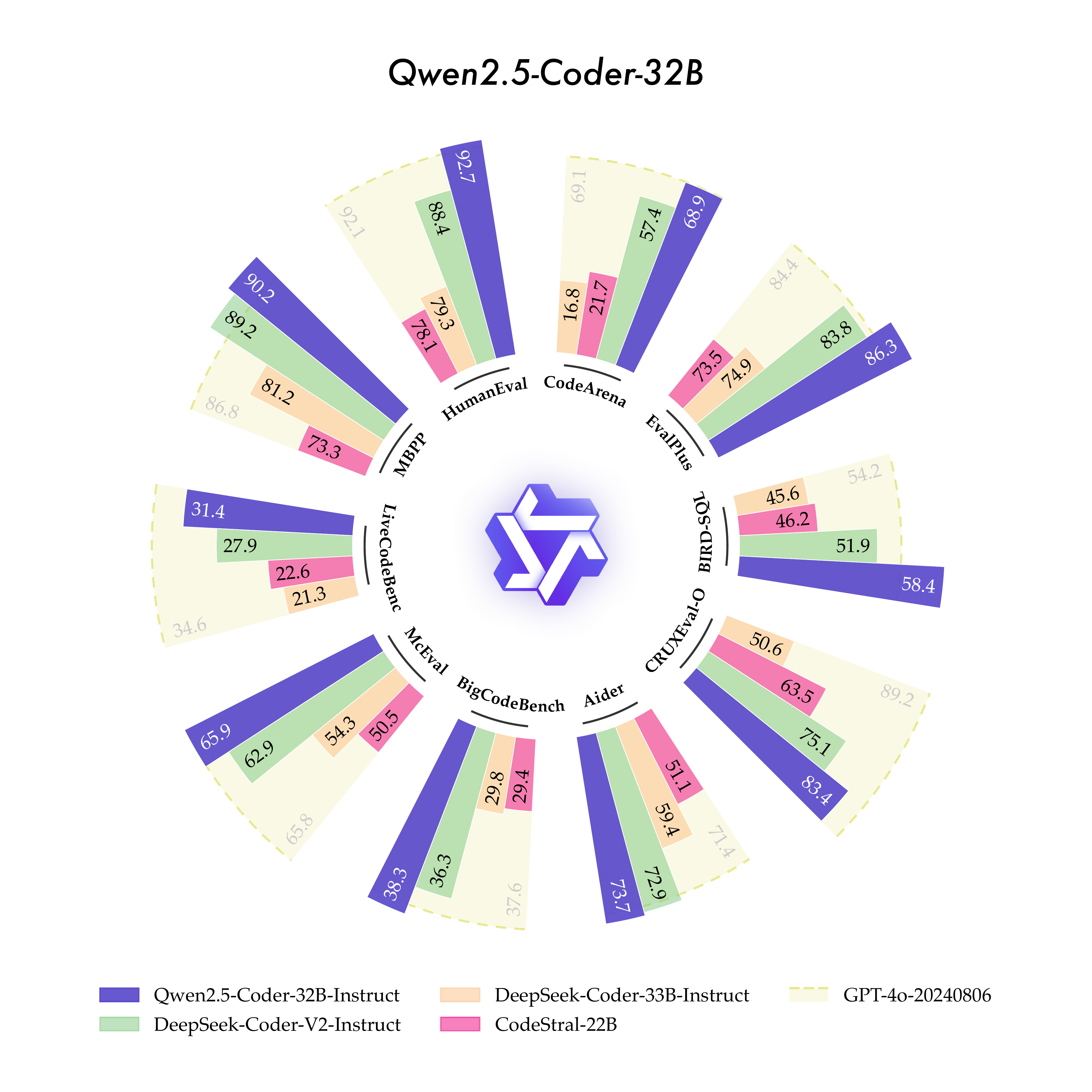}
    \vspace{-1mm}
    \label{fig:intro}
\end{figure}

\begin{spacing}{0.1}
\tableofcontents
\end{spacing}

\newpage

\section{Introduction}
\vspace{-10pt}


With the rapid development of large language models (LLMs)~\citep{gpt3, gpt4, llama2, llama3, mistral, qwen, qwen2, claude3.5, gpt4o}, code-specific language models have garnered significant attention in the community. Built upon pre-trained LLMs, code LLMs such as the StarCoder series~\citep{li2023starcoder, lozhkov2024starcoder2stackv2}, CodeLlama series~\citep{roziere2023codellama}, DeepSeek-Coder series~\citep{guo2024deepseekcoderlargelanguagemodel}, CodeQwen1.5~\citep{codeqwen1.5}, and CodeStral~\citep{codestral}, have demonstrated superior performance in coding evaluations~\citep{chen2021evaluatinglargelanguagemodels, austin2021programsynthesislargelanguage, cassano2022multiplescalableextensibleapproach, jain2024livecodebench,m2rceval,autokaggle,codeeditorbench,tablebench}.
However, in comparison with the recently state-of-the-art proprietary LLMs, Claude-3.5-Sonnet~\citep{claude3.5} and GPT-4o~\citep{gpt4o}, the code LLMs are still falling behind, either open-source or proprietary models.


Building upon our previous work, CodeQwen1.5, we are excited to introduce \textbf{Qwen2.5-Coder}, a new series of language models designed to achieve top-tier performance in coding tasks at various model sizes.
Qwen2.5-Coder models are derived from the Qwen2.5 LLMs, inheriting their advanced architecture and tokenizer. These models are trained on extensive datasets and further fine-tuned on carefully curated instruction datasets specifically designed for coding tasks. 
We are committed to fostering research and innovation in the field of code LLMs, coding agents, and coding assistant applications. Therefore, we release the \textbf{\textit{Powerful}}, \textbf{\textit{Diverse}}, and \textbf{\textit{Practical}} Qwen2.5-Coder series, dedicated to continuously promoting the development of Open CodeLLMs. (1) \textbf{\textit{Powerful}}: Qwen2.5-Coder-32B-Instruct has become the current SOTA open-source code model, matching the coding capabilities of GPT-4o. While demonstrating strong and comprehensive coding abilities, it also possesses good general and mathematical skills. (2) \textbf{\textit{Diverse}}: Qwen2.5-Coder series brings six model sizes, including 0.5B/1.5B/3B/7B/14B/32B. Qwen2.5-Coder has covered six mainstream model sizes to meet the needs of different developers. (3) \textbf{\textit{Practical}}: We explore the practicality of Qwen2.5-Coder in two scenarios, including code assistants and Artifacts, with some examples showcasing the potential applications of Qwen2.5-Coder in real-world scenarios

Significant efforts have been dedicated to constructing a large-scale, coding-specific pretraining dataset comprising over 5.5 trillion tokens. This dataset is sourced from a broad range of public code repositories, such as those on GitHub, as well as large-scale web-crawled data containing code-related texts. We have implemented sophisticated procedures to recall and clean potential code data and filter out low-quality content using weak model based classifiers and scorers. Our approach encompasses both file-level and repository-level pretraining to ensure comprehensive coverage.
To optimize performance and balance coding expertise with general language understanding, we have carefully curated a data mixture that includes code, mathematics, and general texts. 
To transform models into coding assistants for downstream applications, we have developed a well-designed instruction-tuning dataset. This dataset includes a wide range of coding-related problems and solutions, sourced from real-world applications and synthetic data generated by code-focused LLMs, covering a broad spectrum of coding tasks. 

To evaluate the effectiveness of Qwen2.5-Coder, we conducted an extensive evaluation on a suite of popular benchmarks. 
The results highlight Qwen2.5-Coder’s superior code generation capabilities, achieving state-of-the-art performance across more than ten code-focused benchmarks while maintaining robust general and mathematical reasoning abilities. This model outperforms larger code models on a variety of tasks. The release of these models aims to advance code intelligence research and promote widespread adoption in real-world applications, facilitated by permissive licensing.

\section{Model Architecture}

\paragraph{Architecture}
The architecture of \qwencoder is derived directly from Qwen2.5. Table~\ref{tab:training-config} outlines the architecture of \qwencoder across six different model sizes: 0.5B, 1.5B, 3B, 7B, 14B, and 32B parameters. While all sizes share the same architecture in terms of head size, they differ in several other key aspects. With exceptions like the 1.5B model having a larger intermediate size and the 3B model having more layers, most parameters generally increase as the model size scales up. Comparing the 7B and 32B models for instance: the 7B model features a hidden size of 3,584, whereas the 32B model has a hidden size of 5,120. The 7B model uses 28 query heads and 4 key-value heads, while the 32B model uses 40 query heads and 8 key-value heads, reflecting its enhanced capacity. Similarly, the intermediate size scales with model size, being 18,944 for the 7B model and 27,648 for the 32B model. Additionally, smaller models use embedding tying, while larger models do not. Both models have a vocabulary size of 151,646 tokens and are trained on 5.5 trillion tokens.

\paragraph{Tokenization}
\qwencoder inherits the vocabulary from Qwen2.5 but introduces several special tokens to help the model better understand code.
Table \ref{tab:sentinel_tokens} presents an overview of the special tokens added during training to better capture different forms of code data. These tokens serve specific purposes in the code-processing pipeline. For instance, \verb$<|endoftext|>$ marks the end of a text or sequence, while the \verb$<|fim_prefix|>$, \verb$<|fim_middle|>$, and \verb$<|fim_suffix|>$ tokens are used to implement the Fill-in-the-Middle (FIM)  \citep{bavarian2022efficienttraininglanguagemodels} technique, where a model predicts the missing parts of a code block. Additionally, \verb$<|fim_pad|>$ is used for padding during FIM operations. Other tokens include \verb$<|repo_name|>$, which identifies repository names, and \verb$<|file_sep|>$, used as a file separator to better manage repository-level information. These tokens are essential in helping the model learn from diverse code structures and enable it to handle longer and more complex contexts during both file-level and repo-level pretraining.

\begin{table}[htbp]
 \centering
 \begin{tabular}{lcccccc}
 \toprule
 \textbf{Configuration} & \textbf{0.5B} & \textbf{1.5B} & \textbf{3B} & \textbf{7B} & \textbf{14B} & \textbf{32B} \\ \midrule
 Hidden Size & 896 & 1,536 & 2048 & 3,584 & 5120 & 5120 \\
 \#~Layers & 24 & 28 & 36 & 28 & 48 & 64 \\
 \#~Query Heads & 14 & 12 & 16 & 28 & 40 & 40 \\
 \#~KV Heads & 2 & 2 & 2 & 4 & 8 & 8 \\
 Head Size & 128 & 128 & 128 & 128 & 128 & 128 \\
 Intermediate Size & 4,864 & 8,960 & 4,864 & 18,944 & 13824 & 27648 \\
 Embedding Tying & \Checkmark & \Checkmark & \Checkmark & 	\XSolidBrush & 	\XSolidBrush & 	\XSolidBrush \\
 Vocabulary Size & 151,646 & 151,646 & 151,646 & 151,646 & 151,646 & 151,646 \\
 \# Trained Tokens & 5.5T & 5.5T & 5.5T & 5.5T & 5.5T & 5.5T \\
 \bottomrule
 \end{tabular}
 \caption{Architecture of Qwen2.5-Coder.}
 \label{tab:training-config}
\end{table}

\begin{table}[htbp]
    \centering
    \begin{tabular}{lll}
    \toprule
    \textbf{Token} & \textbf{Token ID} & \textbf{Description}\\
    \midrule
    \verb$<|endoftext|>$ & 151643 & end of text/sequence \\ 
    \verb$<|fim_prefix|>$ & 151659 & FIM prefix  \\  
    \verb$<|fim_middle|>$ & 151660 & FIM middle \\
    \verb$<|fim_suffix|>$ & 151661 & FIM suffix \\
    \verb$<|fim_pad|>$ & 151662 & FIM pad \\
    \verb$<|repo_name|>$ & 151663 & repository name \\
    \verb$<|file_sep|>$ & 151664 & file separator  \\
    \bottomrule
    \end{tabular}
    \caption{Overview of the special tokens.}
    \label{tab:sentinel_tokens}
\end{table}

\section{Pre-training}

\subsection{Pretraining Data} Large-scale, high-quality, and diverse data forms the foundation of pre-trained models. To this end, we constructed a dataset named Qwen2.5-Coder-Data. This dataset comprises five key data types: Source Code Data, Text-Code Grounding Data, Synthetic Data, Math Data and Text Data. In this section, we provide a brief overview of the sources and cleaning methods applied to these datasets.

\subsubsection{Data Composition}

\paragraph{Source Code} We collected public repositories from GitHub created before February 2024, spanning 92 programming languages. Similar to StarCoder2 \citep{lozhkov2024starcoder2stackv2} and DS-Coder \citep{guo2024deepseekcoderlargelanguagemodel}, we applied a series of rule-based filtering methods. In addition to raw code, we also collected data from Pull Requests, Commits, Jupyter Notebooks, and Kaggle datasets, all of which were subjected to similar rule-based cleaning techniques.

\begin{figure}[htbp]
    \centering
    \includegraphics[width=0.5\columnwidth]{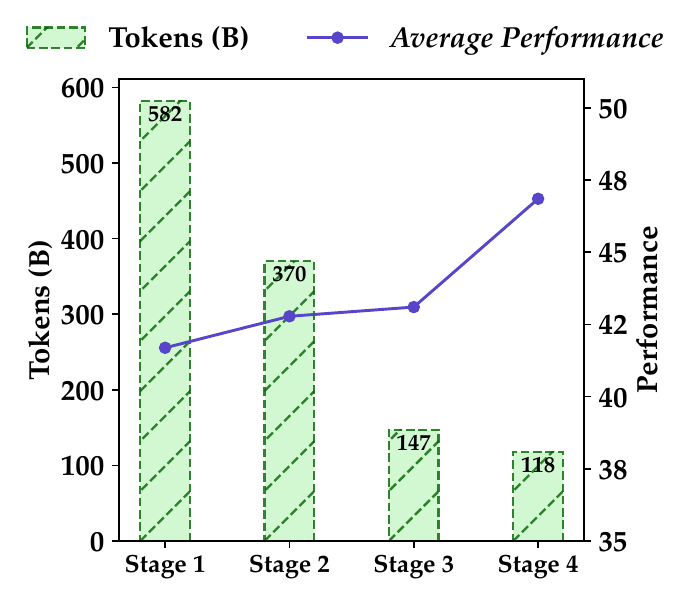}
    \caption{Number of data tokens across different cc-stages, and the validation effectiveness of training \qwencoder using corresponding data.}
    \label{fig:cc-stage-performance}
\end{figure}

\paragraph{Text-Code Grounding Data} We curated a large-scale and high-quality text-code mixed dataset from Common Crawl, which includes code-related documentation, tutorials, blogs, and more. Instead of the conventional URL-based multi-stage recall method, we developed a coarse-to-fine hierarchical filtering approach for raw data. This method offers two key advantages:

\begin{enumerate} \item It enables precise control over each filter's responsibility, ensuring comprehensive handling of each dimension. \item It naturally assigns quality scores to the dataset, with data retained in the final stage being of higher quality, providing valuable insights for quality-driven data mixing. \end{enumerate}

We designed a cleaning pipeline for the Text-Code Grounding Data, where each filter level is built using smaller models, such as fastText. Although we experimented with larger models, they did not yield significant benefits. A likely explanation is that smaller models focus more on surface-level features, avoiding unnecessary semantic complexity.

In \qwencoder, we applied this process iteratively. As shown in Figure \ref{fig:cc-stage-performance}, each iteration resulted in improvement for Qwen2.5-Coder-1.5B. Through 4-stage filtering, the average scores on HumanEval and MBPP increased from 41.6\% to 46.8\% compared to the baseline, demonstrating the value of high-quality Text-Code Grounding Data for code generation.

\paragraph{Synthetic Data} Synthetic data offers a promising way to address the anticipated scarcity of training data. We used CodeQwen1.5, the predecessor of Qwen2.5-Coder, to generate large-scale synthetic datasets. To mitigate the risk of hallucinations during this process, we introduced an executor for validation, ensuring that only executable code was retained.

\paragraph{Math Data} To enhance the mathematical capabilities of \qwencoder, we integrated the pre-training corpus from Qwen2.5-Math into the Qwen2.5-Coder dataset. Importantly, the inclusion of mathematical data did not negatively impact the model's performance on code tasks. For further details on the collection and cleaning process, please refer to the Qwen2.5-Math technical report.

\paragraph{Text Data} Similar to the Math Data, we included high-quality general natural language data from the pre-training corpus of the Qwen2.5 model to preserve \qwencoder's general capabilities. This data had already passed stringent quality checks during the cleaning phase of Qwen2.5's dataset, so no further processing was applied. However, all code segments were removed from the general Text data to avoid overlap with our code data, ensuring the independence of different data sources.

\subsubsection{Data Mixture} Balancing Code, Math, and Text data is crucial for building a foundational model. Although the research community has explored this balance before, there is limited evidence regarding its scalability to large datasets. To address this, we conducted empirical experiments with different ratios of Code, Math, and Text data, designing multiple experiments to identify an optimal combination rapidly. Specifically, as shown in Table \ref{tab:pretrain_domain}, we compared three different Code for Qwen2.5-Coder-7B: Text ratios — 100:0:0, 85:10:5, and 70:20:10.

Interestingly, we found that the 7:2:1 ratio outperformed the others, even surpassing the performance of groups with a higher proportion of code. A possible explanation is that Math and Text data may positively contribute to code performance, but only when their concentration reaches a specific threshold. In future work, we plan to explore more efficient ratio mechanisms and investigate the underlying causes of this phenomenon. Ultimately, we selected a final mixture of 70\% Code, 20\% Text, and 10\% Math. The final training dataset comprises 5.2 trillion tokens.

\begin{table}[h]
\centering
\resizebox{\textwidth}{!}{
\begin{tabular}{ccc|cc|cc|ccc|c}
\toprule
\multicolumn{3}{c|}{\textbf{Token Ratio}} &
\multicolumn{2}{c|}{\textbf{Coding}} & 
\multicolumn{2}{c|}{\textbf{Math}} &
\multicolumn{3}{c|}{\textbf{General}} &
\multirow{2}{*}{\textbf{\textit{Average}}}\\
Code & Text & Math &
Common & BCB & 
MATH & GSM8K &
MMLU & CEval & HellaSwag & \\
\midrule
100 & 0 & 0  &  \fst{49.8} & \fst{40.3} & 10.3 & 23.8 & 42.8 & 35.9 & 58.3 & 31.3 \\
85 & 15 & 5  &  43.3 & 36.2 & 26.1 & 52.5 & 56.8 & 57.1 & 70.0 & 48.9  \\
70 & 20 & 10 &  48.3 & 38.3 & \fst{33.2} & \fst{64.5} & \fst{62.9} & \fst{64.0} & \fst{73.5} & \fst{55.0} \\
\bottomrule
\end{tabular}
}
\caption{The performance of \qwencoder training on different data mixture policy. }
\label{tab:pretrain_domain}
\end{table}

\subsection{Training Policy}

\begin{figure}[htbp]
    \centering
    \includegraphics[width=1.0\columnwidth]{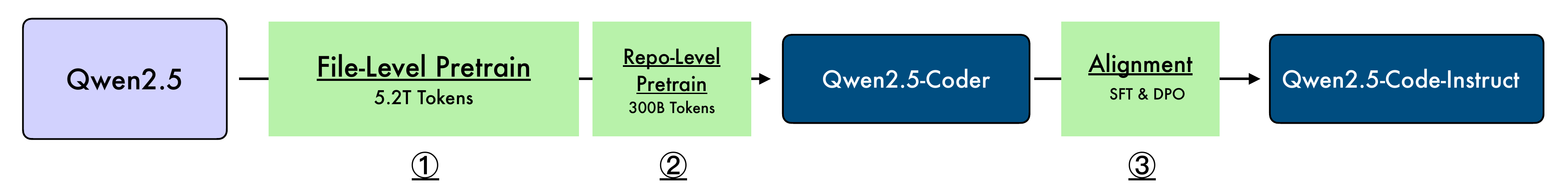}
    \vspace{-5mm}
    \caption{The three-stage training pipeline for \qwencoder.}
    \label{fig:pipeline}
\end{figure}

As shown in \ref{fig:pipeline}, we employed a three-stage training approach to train \qwencoder, including file-level pretraining, repo-level pretraining, and instruction tuning.

\subsubsection{File-Level Pretraining}
\begin{figure}[htbp]
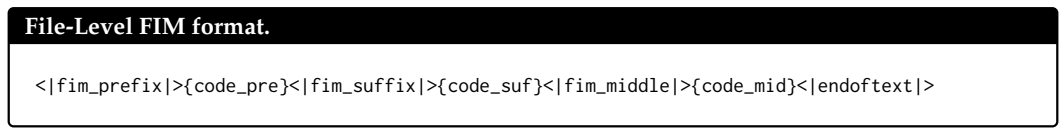

\begin{tcblisting}{listing only, 
    halign=left,
    title=\textbf{\small File-Level FIM format.},
    colbacktitle=black, 
    coltitle=white,
    listing options={basicstyle=\fontsize{8}{13}\selectfont\ttfamily,language=SQL}
}
<|fim_prefix|>{code_pre}<|fim_suffix|>{code_suf}<|fim_middle|>{code_mid}<|endoftext|>
\end{tcblisting}
\caption{File-Level FIM format.}
\label{fig:file-fim}
\end{figure}

File-level pretraining focuses on learning from individual code files. In this stage, the maximum training sequence length is set to 8,192 tokens, covering 5.2T of high-quality data. The training objectives include next token prediction and fill-in-the-middle (FIM) \citep{bavarian2022efficienttraininglanguagemodels}. The specific FIM format is shown in Figure \ref{fig:file-fim}.

\subsubsection{Repo-Level Pretraining}
After file-level pretraining, we turn to repo-level pretraining, aimed at enhancing the model's long-context capabilities. In this stage, the context length is extended from 8,192 tokens to 32,768 tokens, and RoPE’s base frequency is adjusted from 10,000 to 1,000,000. To further leverage the model’s extrapolation potential, we applied the YARN mechanism  \citep{peng2023yarnefficientcontextwindow}, enabling the model to handle sequences up to 131,072 (128K) tokens.

In this stage, we used a large amount of high-quality, long-context code data ($\approx$ 300B) and extended file-level FIM to the repo-level FIM followed by methods described in \citet{lozhkov2024starcoder2stackv2}, with the specific format shown in Figure \ref{fig:repo-fim}.

\begin{figure}[htbp]
\begin{tcblisting}{listing only, 
    halign=left,
    title=\textbf{\small Repo-Level FIM format.},
    colbacktitle=black, 
    coltitle=white,
    listing options={basicstyle=\fontsize{8}{13}\selectfont\ttfamily,language=SQL}
}
<|repo_name|>{repo_name}
<|file_sep|>{file_path1} 
{file_content1}
<|file_sep|>{file_path2} 
{file_content2}
<|file_sep|>{file_path3} 
<|fim_prefix|>{code_pre}<|fim_suffix|>{code_suf}<|fim_middle|>{code_fim}<|endoftext|>
\end{tcblisting}
\caption{Repo-Level FIM format.}
\label{fig:repo-fim}
\end{figure}

\section{Post-training}

\subsection{A Recipe for Instruction Data}

\paragraph{Multilingual Programming Code Identification}
We fine-tune a CodeBERT~\citep{codebert} to perform the language identification model to categorize documents into nearly 100 programming languages. We keep the instruction data of the mainstream programming languages and randomly discard a portion of the instruction data of the long-tail languages. If a given sample contains very little code data or even no code snippets, the sample will possibly be classified into ``No Programming Language'' tag. Since too many instruction samples without code snippets hurt the model performance on code generation tasks (e.g. MultiPL-E, McEval, and MdEval), we remove most of the samples without code snippets to keep the code generation capability of our instruction model.

\paragraph{Instruction Synthesis from GitHub}
For the unsupervised data (code snippets) massively existing in many websites (e.g. GitHub), we try to construct the supervised instruction dataset using LLM. Specifically, we use the LLM to generate the instruction from the code snippets within 1024 tokens and then we use the code LLM to generate the response~\citep{magicoder,unicoder,wavecoder}. Finally, we use the LLM scorer to filter the low-quality ones to obtain the final pair. Given the code snippets of different programming languages, we construct an instruction dataset from the code snippets. To fully unleash the potential of our proposed method, we also include the open-source instruction dataset (e.g. McEval-Instruct for massively multilingual code generation and debugging\footnote{\url{https://huggingface.co/datasets/Multilingual-Multimodal-NLP/McEval-Instruct}}) in the seed instruction dataset. Finally, we combine the instruction data from the GitHub code snippet and open-source instructions for supervised fine-tuning.

\paragraph{Multilingual Code Instruction Data}
To bridge the gap among different programming languages, we propose a multilingual multi-agent collaborative framework to synthesize the multilingual instruction corpora. We introduce language-specific agents, where a set of specialized agents are created and each dedicated to a particular programming language. These agents are initialized with language-specific instruction data derived from the limited existing multilingual instruction corpora. The multilingual data generation process can be split into:
(1) Language-Specific Intelligent Agents: We create a set of specialized agents, each dedicated to a particular programming language. These agents are initialized with language-specific instruction data derived from curated code snippets.
(2) Collaborative Discussion Protocol: Multiple language-specific agents engage in a structured dialogue to formulate new instructions and solutions. This process can result in either enhancing existing language capabilities or generating instructions for a novel programming language.
(3) Adaptive Memory System: Each agent maintains a dynamic memory bank that stores its generation history to avoid generating the similar samples.
(4) Cross-Lingual Discussion: We implement a novel knowledge distillation technique that allows agents to share insights and patterns across language boundaries, fostering a more comprehensive understanding of programming concepts.
(5) Synergy Evaluation Metric: We develop a new metric to quantify the degree of knowledge sharing and synergy between different programming languages within the model.
(6) Adaptive Instruction Generation: The framework includes a mechanism to dynamically generate new instructions based on identified knowledge gaps across languages.

\paragraph{Checklist-based Scoring for Instruction Data}
To completely evaluate the quality of the created instruction pair, we introduce several scoring points for each sample: 
(1) Question\&Answer Consistency: Whether Q\&A are
consistent and correct for fine-tuning. (2) Question\&Answer Relevance: Whether Q\&A are related to the computer field. (3) Question\&Answer Difficulty: Whether Q\&A are sufficiently challenging. (4) Code Exist: Whether the code is provided in question or answer.
(5) Code Correctness: Evaluate whether the provided code is free from syntax errors and logical flaws. (6) Consider factors like proper variable naming, code indentation, and adherence to best practices. (7) Code Clarity: Assess how clear and understandable the code is. Evaluate if it uses meaningful variable names, proper comments, and follows a consistent coding style. (8) Code Comments: Evaluate the presence of
comments and their usefulness in explaining the code's
functionality. (9) Easy to Learn: determine its educational value for a
student whose goal is to learn basic coding concepts. After gaining all scores $(s_{1},\dots,s_{n})$, we can get the final score with $s=w_{1}s_{1}+\dots+w_{n}s_{n}$, where $(w_1,\dots,w_n)$ are a series of pre-defined weights.

\paragraph{A multilingual sandbox for code verification}
To further verify the correctness of the code syntax, we use the code static checking for all extracted code snippets of programming languages (e.g. Python, Java, and C++). We parse the code snippet into the abstract syntax tree and filter out the code snippet, where the parsed nodes in code snippet have parsing errors. We create a multilingual sandbox to support the code static checking for the main programming language.
Further, the multilingual sandbox is a comprehensive platform designed to validate code snippets across multiple programming languages. It automates the process of generating relevant unit tests based on language-specific samples and evaluates whether the provided code snippets can successfully pass these tests. Especially, only the self-contained (e.g. algorithm problems) code snippet will be fed into the multilingual sandbox. The multilingual verification sandbox is mainly comprised of five parts:

\begin{enumerate}
    \item \textbf{Language Support Module:}
    \begin{itemize}
        \item Implements support for multiple languages (e.g., Python, Java, C++, JavaScript)
        \item Maintains language-specific parsing and execution environments
        \item Handles syntax and semantic analysis for each supported language
    \end{itemize}

    \item \textbf{Sample Code Repository:}
    \begin{itemize}
        \item Stores a diverse collection of code samples for each supported language
        \item Organizes samples by language, difficulty level, and programming concepts
        \item Regularly updated and curated by language experts
    \end{itemize}

    \item \textbf{Unit Test Generator:}
    \begin{itemize}
        \item Analyzes sample code to identify key functionalities and edge cases
        \item Automatically generates unit tests based on the expected behavior
        \item Produces test cases covering various input scenarios and expected outputs
    \end{itemize}

    \item \textbf{Code Execution Engine:}
    \begin{itemize}
        \item Provides isolated environments for executing code snippets securely
        \item Supports parallel execution of multiple test cases
        \item Handles resource allocation and timeout mechanisms
    \end{itemize}

    \item \textbf{Result Analyzer:}
    \begin{itemize}
        \item Compares the output of code snippets against expected results from unit tests
        \item Generates detailed reports on test case successes and failures
        \item Provides suggestions for improvements based on failed test cases
    \end{itemize}
\end{enumerate}


\subsection{Training Policy }
\paragraph{Coarse-to-fine Fine-tuning}
We first synthesized tens of millions of low-quality but diverse instruction samples to fine-tune the base model. In the second stage, we adopt millions of high-quality instruction samples to improve the performance of the instruction model with rejection sampling and supervised fine-tuning. For the same query, we use the LLM to generate multiple candidates and then use the LLM to score the best one for supervised fine-tuning.

\paragraph{Mixed Tuning}
Since most instruction data have a short length, we construct the instruction pair with the FIM format to keep the long context capability of the base model. Inspired by programming language syntax rules and user habits in practical scenarios, we leverage the \texttt{tree-sitter-languages}\footnote{\url{https://pypi.org/project/tree-sitter-languages/}} to parse the code snippets and extract the basic logic blocks as the middle code to infill. For example, the abstract syntax tree (AST) represents the structure of Python code in a tree format, where each node in the tree represents a construct occurring in the source code. The tree's hierarchical nature reflects the syntactic nesting of constructs in the code and includes various elements such as expressions, statements, and functions. By traversing and manipulating the AST, we can randomly extract the nodes of multiple levels and use the code context of the same file to uncover the masked node. Finally, we optimize the instruction model with a majority of standard SFT data and a small part of FIM instruction samples. 

\paragraph{Direct Preference Optimization for Code}
After obtaining the SFT model, we further align the Qwen2.5-Coder with the help of offline direct preference optimization (DPO)~\citep{rafailov2023direct}. Given that human feedback is highly labor-intensive, we use a multilingual code sandbox to provide code execution feedback, while an LLM is utilized for human judgment feedback. For the algorithm-like and self-contained code snippets, we generate the test cases to check the correctness of the code as the code execution feedback, including Python, Java, and other languages. For other complex code snippets, we use LLM-as-a-judge~\citep{zheng2023judging} to decide which code snippet is better. Further, we combine the code DPO data and common data for offline DPO training.

\section{Decontamination}
To ensure that \qwencoder does not produce inflated results due to test set leakage, we performed decontamination on all data, including both pre-training and post-training datasets. We removed key datasets such as HumanEval, MBPP, GSM8K, and MATH. The filtering was done using a 10-gram overlap method, where any training data with a 10-gram word-level overlap with the test data was removed.

\section{Evaluation on Base Models}
For the base model, we conducted a comprehensive and fair evaluation in six key aspects, including code generation, code completion, code reasoning, mathematical reasoning, general natural language understanding and long-context modeling. To ensure the reproducibility of all results, we made all evaluation codes publicly available\footnote{\url{https://github.com/QwenLM/Qwen2.5-Coder}}. For comparing models, we chose the most popular and powerful open source language models, including the StarCoder2 and DeepSeek-Coder series. Below is the list of artifacts used in the evaluation for this section.

\begin{table}[htbp]
 \small
 \centering
 \begin{tabular}{l|l}
 \toprule
 \textbf{Artifact} & \textbf{Public link} \\
 \midrule
 Qwen2.5-Coder-0.5B & \url{https://hf.co/Qwen/Qwen2.5-Coder-0.5B} \\
 Qwen2.5-Coder-1.5B & \url{https://hf.co/Qwen/Qwen2.5-Coder-1.5B} \\
 Qwen2.5-Coder-3B & \url{https://hf.co/Qwen/Qwen2.5-Coder-3B} \\
 Qwen2.5-Coder-7B & \url{https://hf.co/Qwen/Qwen2.5-Coder-7B} \\
 Qwen2.5-Coder-14B & \url{https://hf.co/Qwen/Qwen2.5-Coder-14B} \\
 Qwen2.5-Coder-32B & \url{https://hf.co/Qwen/Qwen2.5-Coder-32B} \\
 \midrule
 CodeQwen1.5-7B & \url{https://hf.co/Qwen/CodeQwen1.5-7B} \\
 StarCoder2-3B & \url{https://hf.co/bigcode/starcoder2-3b} \\
 StarCoder2-7B & \url{https://hf.co/bigcode/starcoder2-7b} \\
 StarCoder2-15B & \url{https://hf.co/bigcode/starcoder2-15b} \\
 DS-Coder-1.3B-Base & \url{https://hf.co/deepseek-ai/deepseek-coder-1.3b-base} \\
 DS-Coder-6.7B-Base & \url{https://hf.co/deepseek-ai/deepseek-coder-6.7b-base} \\
 DS-Coder-33B-Base & \url{https://hf.co/deepseek-ai/deepseek-coder-33b-base} \\
 DS-Coder-V2-Lite-Base & \url{https://hf.co/deepseek-ai/DeepSeek-Coder-V2-Lite-Base} \\
 DS-Coder-V2-Base & \url{https://hf.co/deepseek-ai/DeepSeek-Coder-V2-Base} \\
 \bottomrule
 \end{tabular}
 \caption{All artifacts released and used in this section.}
 \label{tab:artifacts}
\end{table}

\subsection{Code Generation}

\begin{table}[h]
\centering
\resizebox{\textwidth}{!}{
\begin{tabular}{lr|cc|ccc|cc}
\toprule
\multirow{2}{*}{\textbf{Model}} & 
\multirow{2}{*}{\textbf{Size}} & 
\multicolumn{2}{c|}{\textbf{HumanEval}} & 
\multicolumn{3}{c|}{\textbf{MBPP}} &
\multicolumn{2}{c}{\textbf{BigCodeBench}} \\
& & 
\textit{HE} & \textit{HE+} & 
\textit{MBPP} & \textit{MBPP+} & \textit{3-shot} &
\textit{Full} & \textit{Hard}  \\
\midrule
\multicolumn{9}{c}{\textbf{0.5B+ Models}} \\
\midrule
\graybg{}\textbf{Qwen2.5-Coder-0.5B} & 0.5B & \fst{28.0} & \fst{23.8} & \fst{52.9} & \fst{47.1} & \fst{40.4} & \fst{16.1} & \fst{4.7}\\
\midrule
\multicolumn{9}{c}{\textbf{1B+ Models}} \\
\midrule
DS-Coder-1.3B & 1.3B & 34.8 & 26.8 & 55.6 & 46.9 & 46.2 & 26.1 & 3.4 \\
\graybg{}\textbf{Qwen2.5-Coder-1.5B} & 1.5B & \fst{43.9} & \fst{36.6} & \fst{69.2} & \fst{58.6} & \fst{59.2} & \fst{34.6} & \fst{9.5}\\
\midrule
\multicolumn{9}{c}{\textbf{3B+ Models}} \\
\midrule
StarCoder2-3B & 3B & 31.7 & 27.4 & 60.2 & 49.1 & 47.4 & 21.4 & 4.7 \\
\graybg{}\textbf{Qwen2.5-Coder-3B} & 3B & \fst{52.4} & \fst{42.7} & \fst{72.2} & \fst{61.4} & \fst{65.2} & \fst{41.1} & \fst{11.5}\\
\midrule
\multicolumn{9}{c}{\textbf{6B+ Models}} \\
\midrule
StarCoder2-7B & 7B & 35.4 & 29.9 & 54.4 & 45.6 & 51.8 & 27.7 & 8.8 \\
DS-Coder-6.7B-Base & 6.7B & 47.6 & 39.6 & 70.2 & 56.6 & 60.6 & 41.1 & 11.5 \\
DS-Coder-V2-Lite-Base & 2.4/16B & 40.9 & 34.1 & 71.9 & 59.4 & 62.6 & 30.6 & 8.1 \\
CodeQwen1.5-7B & 7B & 51.8 & 45.7 & 72.2 & 60.2 & 61.8 & 45.6 & 15.5 \\
\graybg{}\textbf{Qwen2.5-Coder-7B} & 7B & \fst{61.6} & \fst{53.0} & \fst{76.9} & \fst{62.9} & \fst{68.8} & \fst{45.8} & \fst{16.2} \\
\midrule
\multicolumn{9}{c}{\textbf{14B+ Models}} \\
\midrule
StarCoder2-15B & 15B & 46.3 & 37.8 & 66.2 & 53.1 & 57.0 & 38.4 & 12.2 \\
\graybg{}\textbf{Qwen2.5-Coder-14B} & 14B & \fst{64.0} & \fst{57.9} & \fst{81.0} & \fst{66.7} & \fst{71.4} & \fst{51.8} & \fst{22.3} \\
\midrule
\multicolumn{9}{c}{\textbf{20B+ Models}} \\
\midrule
DS-Coder-33B-Base &  33B &  54.9 & 47.6 & 74.2 & 60.7 & 66.0 & 49.1 & 20.3 \\
DS-Coder-V2-Base &  21/236B & 50.0 &  43.3 &  82.5 &  65.7 & 71.2 & 48.7 & 21.6 \\
\graybg{}\textbf{Qwen2.5-Coder-32B} & 32B & \fst{65.9} & \fst{60.4} & \fst{83.0} & \fst{68.2} & \fst{76.4} & \fst{53.6} & \fst{26.4} \\

\bottomrule
\end{tabular}
}
\caption{Performance of various models on HumanEval, MBPP and the ``complete'' task of BigCodeBench. }
\label{tab:base-codegen}
\end{table}

\begin{table}[h]
\centering
\resizebox{\textwidth}{!}{
\begin{tabular}{lr|cccccccc|c}
\toprule
\textbf{Model} & \textbf{Size} & 
Python & C++ & Java & PHP & TS & C\# & Bash & JS & \textbf{Average}\\
\midrule
\multicolumn{11}{c}{\textbf{0.5B+ Models}} \\
\midrule
\graybg{}\textbf{Qwen2.5-Coder-0.5B} & 0.5B & \fst{28.0} & \fst{25.5} & \fst{22.8} & \fst{23.6} & \fst{30.8} & \fst{31.0} & \fst{7.0} & \fst{29.2} & \fst{24.7}\\
\midrule
\multicolumn{11}{c}{\textbf{1B+ Models}} \\
\midrule
DS-Coder-1.3B-Base & 1.3B & 34.8 & 31.1 & 32.3 & 24.2 & 28.9 & 36.7 & 10.1 & 28.6 & 28.3\\
\graybg{}\textbf{Qwen2.5-Coder-1.5B} & 1.5B & \fst{42.1} & \fst{42.9} & \fst{38.6} & \fst{41.0} & \fst{49.1} & \fst{46.2} & \fst{20.3} & \fst{49.1} & \fst{41.1}\\
\midrule
\multicolumn{11}{c}{\textbf{3B+ Models}} \\
\midrule
StarCoder2-3B & 3B & 31.7 & 30.4 & 29.8 & 32.9 & 39.6 & 34.8 & 13.9 & 35.4 & 31.1\\
\graybg{}\textbf{Qwen2.5-Coder-3B} & 3B & \fst{52.4} & \fst{52.8} & \fst{44.9} & \fst{49.1} & \fst{55.4} & \fst{51.3} & \fst{24.7} & \fst{53.4} & \fst{48.0}\\
\midrule
\multicolumn{11}{c}{\textbf{6B+ Models}} \\
\midrule
StarCoder2-7B & 7B & 35.4 & 40.4 & 38.0 & 30.4 & 34.0 & 46.2 & 13.9 & 36.0 & 34.3\\
DS-Coder-6.7B-Base & 6.7B & 49.4 & 50.3 & 43.0 & 38.5 & 49.7 & 50.0 & 28.5 & 48.4 & 44.7\\
DS-Coder-V2-Lite-Base & 2.4/16B & 40.9 & 45.9 & 34.8 & 47.2 & 48.4 & 41.7 & 19.6 & 44.7 & 40.4\\
CodeQwen1.5-7B & 7B & 51.8 & 52.2 & 42.4 & 46.6 & 52.2 & 55.7 & 36.7 & 49.7 & 48.4\\
\graybg{}\textbf{Qwen2.5-Coder-7B} & 7B & \fst{61.6} & \fst{62.1} & \fst{53.2} & \fst{59.0} & \fst{64.2} & \fst{60.8} & \fst{38.6} & \fst{60.3} & \fst{57.5}\\
\midrule
\multicolumn{11}{c}{\textbf{14B+ Models}} \\
\midrule
StarCoder2-15B & 15B & 46.3 & 47.2 & 46.2 & 39.1 & 42.1 & 53.2 & 15.8 & 43.5 & 41.7\\
\graybg{}\textbf{Qwen2.5-Coder-14B} & 14B & \fst{64.0} & \fst{69.6} & \fst{46.8} & \fst{64.6} & \fst{69.2} & \fst{63.3} & \fst{39.9} & \fst{61.5} & \fst{59.9}\\	 	 	 
\midrule
\multicolumn{11}{c}{\textbf{20B+ Models}} \\
\midrule
DS-Coder-33B-Base & 33B & 56.1 & 58.4 & 51.9 & 44.1 & 52.8 & 51.3 & 32.3 & 55.3 & 50.3\\
DS-Coder-V2-Base & 21/236B & 50.0 & 59.6 & 50.0 & 55.3 & 58.5 & 45.6 & 36.1 & 59.6 & 51.8\\
\graybg{}\textbf{Qwen2.5-Coder-32B} & 32B & \fst{65.9} & \fst{68.3} & \fst{70.9} & \fst{64.6} & \fst{66.0} & \fst{68.4} & \fst{39.9} & \fst{67.1} & \fst{63.9}\\	 	 	
\bottomrule
\end{tabular}
}
\caption{Performance of different models on MultiPL-E.}
\label{tab:base-multiple}
\end{table}

\paragraph{HumanEval and MBPP}
Code generation serves as a fundamental capability for code models to handle more complex tasks. We selected two popular code generation benchmarks to evaluate \qwencoder, namely HumanEval \citep{chen2021evaluatinglargelanguagemodels} and MBPP \citep{austin2021programsynthesislargelanguage}. HumanEval consists of 164 manually written programming tasks, each providing a Python function signature and a docstring as input to the model. MBPP, on the other hand, comprises 974 programming problems created by crowdsource contributors. Each problem includes a problem statement (i.e., a docstring), a function signature, and three test cases.

To further ensure accurate evaluation, EvalPlus \citep{evalplus} extends HumanEval into HumanEval+ by adding 80 times more unique test cases and correcting inaccurate ground-truth solutions in HumanEval. Similarly, MBPP+ offers 35 times more test cases than the original MBPP.

Additionally, we should notice that MBPP 3-shot is particularly suitable for monitoring model convergence during training. Early in the convergence process, the model tends to be unstable, causing significant fluctuation in metrics, and simple 3-shot examples effectively mitigate it. Therefore, we also report the results of MBPP 3-shot performance.

As shown in Table \ref{tab:base-codegen}, \qwencoder have shown impressive performance in basic code generation, achieving state-of-the-art results among open-source models of the same size and surpassing even larger models. In particular, Qwen2.5-Coder-7B outperforms the previous best dense model, DS-Coder-33B, across all five metrics.

\paragraph{BigCodeBench-Complete}
BigCodeBench \citep{zhuo2024bigcodebenchbenchmarkingcodegeneration} is a recent and more challenging benchmark for code generation, primarily aimed at evaluating the ability of tool-use and complex instruction following. The base model generates the expected code through a completion mode, given a function signature and documentation, which is referred to as BigCodeBench-Complete. It consists of two subsets: the full set and the hard set. Compared to HumanEval and MBPP, BigCodeBench is suited for out-of-distribution (OOD) evaluation.

Table \ref{tab:base-codegen} illustrates that \qwencoder continues to show strong performance on BigCodeBench-Complete, underscoring the model's generalization potential.

\paragraph{Multi-Programming Language}
The evaluations mentioned above focus on the Python language. However, we expect a strong code model to be not only proficient in Python but also versatile across multiple programming languages to meet the complex and evolving demands of software development. To more comprehensively evaluate \qwencoder's proficiency in handling multiple programming languages, we selected the MultiPL-E \citep{cassano2022multiplescalableextensibleapproach} and chose to evaluate eight mainstream languages from this benchmark, including Python, C++, Java, PHP, TypeScript, C\#, Bash and JavaScript.

As shown in the table \ref{tab:base-multiple}, \qwencoder also achieved state-of-the-art results in the multi-programming language evaluation, with its capabilities well-balanced across various languages. It scored over 60\% in five out of the eight languages.

\subsection{Code Completion}
\begin{figure}[htbp]
    \centering
    \includegraphics[width=0.9\columnwidth]{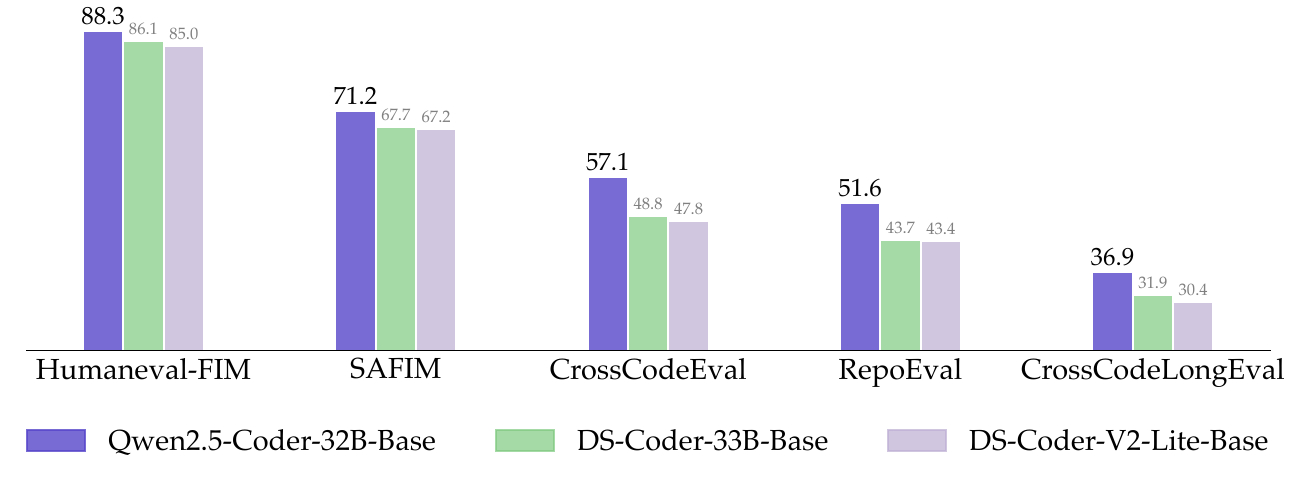}
    \vspace{-1mm}
    \caption{The code completion performance of competitive models on five benchmarks, Humaneval-FIM, SAFIM, CrossCodeEval, RepoEval, CrossCodeLongEval.}
    \label{fig:fim_score_overall}
\end{figure}

Many developer aid tools rely on the capability to autocomplete code based on preceding and succeeding code snippets. \qwencoder utilizes the Fill-In-the-Middle (FIM) training strategy, as introduced in \cite{bavarian2022efficienttraininglanguagemodels}, enabling the model to generate code that is contextually coherent. To assess its code completion proficiency, we utilize the HumanEval-FIM benchmark~\citep{allal2023santacoder}, CrossCodeEval~\citep{ding2024crosscodeeval},  CrossCodeLongEval~\citep{wu2024repoformer}, RepoEval~\citep{zhang2023repocoder} and SAFIM~\citep{gong2024evaluation}. Figure \ref{fig:fim_score_overall} shows the overall evaluation results of Qwen2.5-Coder-32B on different code completion benchmarks.

Humaneval-FIM benchmark challenges the model to accurately predict missing sections of code within tasks derived from Humaneval. We use the single-line infilling settings across Python, Java, and JavaScript, focusing on predicting a single line of code within given contexts. Performance was measured using the Exact Match metric, which determines the proportion of the first generated code line that precisely match the ground truth. The table \ref{tab:base-fim} illustrates that \qwencoder surpasses alternative models concerning model size. Specifically, Qwen2.5-Coder-1.5B achieves an average performance improvement of 3.7\%, rivaling the majority of models exceeding 6 billion parameters. Moreover, Qwen2.5-Coder-7B stands as the leading model among those over 6 billion parameters, matching the performance of the formidable 33 billion parameter model, DS-Coder-33B-Base. Notably, we excluded DS-Coder-v2-236B from comparison due to its design focus not being on code completion tasks.

\begin{table}[h]
\centering
\begin{tabular}{lr|cccc}
\toprule
\multirow{2}{*}{\textbf{Model}} & 
\multirow{2}{*}{\textbf{Size}} & 
\multicolumn{4}{c}{\textbf{Humaneval-FIM}}  \\
& &  
\textit{Python} & \textit{Java} & \textit{JavaScript} &  \textit{Average}$^*$ \\
\midrule
\multicolumn{6}{c}{\textbf{0.5B+ Models}} \\
\midrule
\graybg{}\textbf{Qwen2.5-Coder-0.5B} & 0.5B & \fst{70.3} & \fst{78.1} & \fst{81.2} & \fst{77.7} \\
\midrule
\multicolumn{6}{c}{\textbf{1B+ Models}} \\
\midrule
DS-Coder-1.3B-Base & 1.3B & 72.8 & 84.3	& 81.7 & 80.7 \\
\graybg{}\textbf{Qwen2.5-Coder-1.5B} & 1.5B & \fst{77.0} & \fst{85.6} & \fst{85.0} & \fst{83.5} \\
\midrule
\multicolumn{6}{c}{\textbf{3B+ Models}} \\
\midrule
StarCoder2-3B & 3B & 70.9 &	84.4 & 81.8 & 80.4 \\
\graybg{}\textbf{Qwen2.5-Coder-3B} & 3B & \fst{78.7} & \fst{88.0} & \fst{87.4} & \fst{85.7} \\
\midrule
\multicolumn{6}{c}{\textbf{6B+ Models}} \\
\midrule
StarCoder2-7B & 7B & 70.8 &	86.0 & 84.4 & 82.0 \\
DS-Coder-6.7B-Base & 6.7B & 78.1 & 87.4 & 84.1 & 84.0 \\
DS-Coder-V2-Lite-Base & 2.4/16B & 78.7 & 87.8 & 85.9 & 85.0 \\
CodeQwen1.5-7B & 7B & 75.8	& 85.7 & 85.0 & 83.3 \\
\graybg{}\textbf{Qwen2.5-Coder-7B} & 7B & \fst{79.7} & \fst{88.5} & \fst{87.6} & \fst{86.2} \\
\midrule
\multicolumn{6}{c}{\textbf{14B+ Models}} \\
\midrule
StarCoder2-15B & 15B & 74.2 & 85.2 & 84.6 & 82.6 \\
\graybg{}\textbf{Qwen2.5-Coder-14B} & 14B & \fst{80.5} & \fst{91.0} & \fst{88.5} & \fst{87.7} \\
\midrule
\multicolumn{6}{c}{\textbf{20B+ Models}} \\
\midrule
 CodeStral-22B & 22B &  76.7 & 	82.5	&  86.0 &  82.7 \\
 DS-Coder-33B-Base & 33B &  80.1 & 	89.0	&  86.8 &  86.2 \\
\graybg{}\textbf{Qwen2.5-Coder-32B} & 32B & \fst{81.5} & \fst{91.0} & \fst{89.4} & \fst{88.3} \\
\bottomrule
\end{tabular}
\caption{Performance of different approaches on the Humaneval-FIM Tasks. ${}^*$\textit{Average} refers to a weighted mean calculated based on the number of samples for each language.}
\label{tab:base-fim}
\end{table}

In real-world scenarios, code completion often depends on accessing cross-file context and dependencies. CrossCodeEval is a benchmark that requires a deep understanding of this cross-file context to accurately complete the code. In our evaluation, we set a maximum sequence length of 8192 tokens, designate a maximum output length of 50 tokens, and impose a limit of 2048 tokens for the cross-file context. For the cross-file context, we use the official BM25 search results provided by \cite{ding2024crosscodeeval}. We evaluate performance using Exact Match (EM) and Edit Similarity (ES) metrics. Table \ref{tab:base-cceval} shows that the Qwen2.5-Coder-32B achieves state-of-the-art performance with a 3.7\% improvement. Qwen2.5-Coder outperforms all the models with a comparable model size. Meanwhile, Qwen2.5-Coder-7B has a comparable performance with other models exceeding 20 billion parameters.
 
\begin{table}[h]
\centering
\small
\begin{tabular}{l| c c c c c c c c c c}
\toprule
\multirow{2}{*}{\textbf{Model}} &\multicolumn{2}{c}{Python}&\multicolumn{2}{c}{Java}&\multicolumn{2}{c}{TypeScript}&\multicolumn{2}{c}{C\#} &\multicolumn{2}{c}{Average} \\
\cmidrule(lr){2-3}\cmidrule(lr){4-5}\cmidrule(lr){6-7}\cmidrule(lr){8-9}\cmidrule(lr){10-11}&\textit{EM}&\textit{ES}&\textit{EM}&\textit{ES}&\textit{EM}&\textit{ES}&\textit{EM}&\textit{ES}&\textit{EM}&\textit{ES}\\
\midrule
\multicolumn{11}{c}{\textbf{0.5B+ Models}} \\
\midrule
\graybg{}\textbf{Qwen2.5-Coder-0.5B} & \fst{22.7} & \fst{66.2} & \fst{21.7} & \fst{66.8} & \fst{21.9} & \fst{67.2} & \fst{32.1} & \fst{75.4} & \fst{24.6} & \fst{68.9} \\
\midrule
\multicolumn{11}{c}{\textbf{1B+ Models}} \\
\midrule
DS-Coder-1.3B-Base & 33.4 & 72.6 & 34.9 & 74.5 & 36.7 & 76.4 & 46.6 & 83.5 & 37.9 & 76.8 \\
\graybg{}\textbf{Qwen2.5-Coder-1.5B} & \fst{35.5} & \fst{74.3} & \fst{37.9} & \fst{76.5} & \fst{37.6} & \fst{77.4} & \fst{49.8} & \fst{84.5} & \fst{40.2} & \fst{78.2}
 \\
\midrule
\multicolumn{11}{c}{\textbf{3B+ Models}} \\
\midrule
StarCoder2-3B & 11.0 & 62.7 & 11.6 & 69.7 & 8.8 & 75.8 & 8.2 & 71.2 & 9.9 & 69.8 \\
\graybg{}\textbf{Qwen2.5-Coder-3B} & \fst{38.4} & \fst{76.1} & \fst{42.8} & \fst{79.8} & \fst{41.6} & \fst{80.5} & \fst{56.7} & \fst{87.1} & \fst{44.9} & \fst{80.9} \\
\midrule
\multicolumn{11}{c}{\textbf{6B+ Models}} \\
\midrule
StarCoder2-7B & 10.9 & 63.1 & 8.3 & 71.0 & 6.7 & 76.8 & 7.3 & 72.1 & 8.3 & 70.8 \\
DS-Coder-6.7B-Base & 41.1 & 79.2 & 39.9 & 80.1 & 46.3 & 82.4 & 55.0 & 86.9 & 45.6 & 82.1 \\
DS-Coder-V2-Lite-Base & 41.8 & 78.3 & 46.1 & 81.2 & 44.6 & 81.4 & 58.7 & 87.9 & 47.8 & 82.2 \\
CodeQwen1.5-7B & 40.7 & 77.8 & 47.0 & 81.6 & 45.8 & 82.2 & 59.7 & 87.6 & 48.3 & 82.3 \\
\graybg{}\textbf{Qwen2.5-Coder-7B} & \fst{42.4} & \fst{78.6} & \fst{48.1} & \fst{82.6} & \fst{46.8} & \fst{83.4} & \fst{59.7} & \fst{87.9} & \fst{49.3} & \fst{83.1} \\
\midrule
\multicolumn{11}{c}{\textbf{14B+ Models}} \\
\midrule
StarCoder2-15B & 28.2 & 70.5 & 26.7 & 71.0 & 24.7 & 76.3 & 25.2 & 74.2 & 26.2 & 73.0 \\
\graybg{}\textbf{Qwen2.5-Coder-14B} &\fst{47.7} & \fst{81.7} & \fst{54.7} & \fst{85.7} & \fst{52.9} & \fst{86.0} & \fst{66.4} & \fst{91.1} & \fst{55.4} & \fst{86.1} \\
\midrule
\multicolumn{11}{c}{\textbf{20B+ Models}} \\
\midrule
 CodeStral-22B & \fst{49.3} & \fst{82.7} & 44.1 & 71.1 & 51.0 & 85.0 & 53.7 & 83.6 & 49.5 & 80.6 \\
 DS-Coder-33B-Base & 44.2 & 80.4 & 46.5 & 82.7 & 49.2 & 84.0 & 55.2 & 87.8 & 48.8 & 83.7 \\
\graybg{}\textbf{Qwen2.5-Coder-32B} & 49.2 & 82.1 & \fst{56.4} & \fst{86.6} & \fst{54.9} & \fst{87.0} & \fst{68.0} & \fst{91.6} & \fst{57.1} & \fst{86.8} \\
\bottomrule
\end{tabular}
\caption{Performance of different approaches on the CrossCodeEval Tasks.}
\label{tab:base-cceval}
\end{table}

CrossCodeLongEval is a long context benchmark on cross file code completion tasks. In our evaluation, we set a maximum sequence length of 8192 tokens and set the maximum output as 256 tokens for function completion and 50 tokens for other tasks. The cross-file context is truncated to 2048 tokens. For the cross-file context, we use the official BM25 search results provided by \cite{wu2024repoformer}. We evaluate performance using Exact Match (EM) and Edit Similarity (ES) metrics. Qwen2.5-Coder-32B achieves state-of-the-art performance, as detailed in Table \ref{tab:base-cclongeval}.  The Qwen2.5-Coder series surpasses all other models of a similar size. All models demonstrate low Exact Match (EM) results on function completion tasks, likely due to the complexity of generating multi-line code snippets that are challenging to match precisely.

\begin{table}[h]
\centering
\begin{tabular}{l| c c c c c c}
\toprule
\multirow{2}{*}{\textbf{Model}} &\multicolumn{2}{c}{Chunk Completion}&\multicolumn{2}{c}{Function completion} &\multicolumn{2}{c}{Average} \\
\cmidrule(lr){2-3}\cmidrule(lr){4-5}\cmidrule(lr){6-7}&\textit{EM}&\textit{ES}&\textit{EM}&\textit{ES}&\textit{EM}&\textit{ES}\\
\midrule
\multicolumn{7}{c}{\textbf{0.5B+ Models}} \\
\midrule
\graybg{}\textbf{Qwen2.5-Coder-0.5B} & \fst{29.8} & \fst{64.2} & \fst{9.5} & \fst{38.0} & \fst{19.7} & \fst{51.1} \\
\midrule
\multicolumn{7}{c}{\textbf{1B+ Models}} \\
\midrule
DS-Coder-1.3B-Base & 40.6 & 71.9 & 9.6 & 39.4 & 25.1 & 55.7 \\
\graybg{}\textbf{Qwen2.5-Coder-1.5B} & \fst{44.2} & \fst{73.9} & \fst{12.4} & \fst{44.4} & \fst{28.3} & \fst{59.2}
 \\
\midrule
\multicolumn{7}{c}{\textbf{3B+ Models}} \\
\midrule
StarCoder2-3B & 18.5 & 62.0 & 10.2 & 39.2 & 14.3 & 50.6 \\
\graybg{}\textbf{Qwen2.5-Coder-3B} & \fst{46.6} & \fst{76.1} & \fst{13.5} & \fst{46.4} & \fst{30.0} & \fst{61.3} \\
\midrule
\multicolumn{7}{c}{\textbf{6B+ Models}} \\
\midrule
StarCoder2-7B & 19.4 & 63.6 & 10.2 & 40.0 & 14.8 & 51.8 \\
DS-Coder-6.7B-Base & 48.4 & 78.2 & 10.7 & 42.4 & 29.6 & 60.3 \\
DS-Coder-V2-Lite-Base & 49.5 & 77.1 & 11.4 & 43.1 & 30.4 & 60.1 \\
CodeQwen1.5-7B & 48.2 & 77.5 & 6.4 & 30.6 & 27.3 & 54.1 \\
\graybg{}\textbf{Qwen2.5-Coder-7B} & \fst{52.4} & \fst{79.3} & \fst{14.4} & \fst{48.4} & \fst{33.4} & \fst{63.8} \\
\midrule
\multicolumn{7}{c}{\textbf{14B+ Models}} \\
\midrule
StarCoder2-15B & 21.3 & 53.7 & 7.8 & 30.5 & 14.6 & 42.1 \\
\graybg{}\textbf{Qwen2.5-Coder-14B} & \fst{56.9} & \fst{81.8} & \fst{15.4} & \fst{49.8} & \fst{36.1} & \fst{65.8} \\
\midrule
\multicolumn{7}{c}{\textbf{20B+ Models}} \\
\midrule
 CodeStral-22B &56.7 & 81.8 & 10.5 & 37.8 & 33.6 & 59.8 \\
 DS-Coder-33B-Base & 52.0 & 79.9 & 11.9 & 44.3 & 32.0 & 62.1 \\
\graybg{}\textbf{Qwen2.5-Coder-32B} & \fst{57.3} & \fst{82.1} & \fst{16.4} & \fst{50.8} & \fst{36.9} & \fst{66.4} \\
\bottomrule
\end{tabular}
\caption{Performance of different approaches on the CrossCodeLongEval Tasks.}
\label{tab:base-cclongeval}
\end{table}


RepoEval is a benchmark designed to evaluate repository-level code completion capabilities across three granularities: line, API invocation, and function body completion. In our evaluation, we set a maximum sequence length of 8192 tokens, set the maximum output as 256 tokens for function completion and 50 tokens for other tasks, and impose a limit of 2048 tokens for the cross-file context. Besides, we utilize the official sparse retriever~\citep{sparse} to extract the cross-file context. We evaluate performance using Exact Match (EM) and Edit Similarity (ES) metrics. As shown in Table \ref{tab:base-repoeval}, Qwen2.5-Coder-32B achieves state-of-the-art performance with an average improvement of 7.9\% EM and 4.2\% ES compared to DS-Coder-33B-Base. Furthermore, Qwen2.5-Coder-14B and Qwen2.5-Coder-7B achieve comparable performance to models with more than 20B parameters, while maintaining state-of-the-art results among models of similar size.

\begin{table}[h]
\centering
\small
\begin{tabular}{l| c c c c c c c c c c}
\toprule
\multirow{2}{*}{\textbf{Model}} &\multicolumn{2}{c}{Line}&\multicolumn{2}{c}{Function}&\multicolumn{2}{c}{API}&\multicolumn{2}{c}{Average} \\
\cmidrule(lr){2-3}\cmidrule(lr){4-5}\cmidrule(lr){6-7}\cmidrule(lr){8-9}\cmidrule(lr){10-11}&\textit{EM}&\textit{ES}&\textit{EM}&\textit{ES}&\textit{EM}&\textit{ES}&\textit{EM}&\textit{ES}\\
\midrule
\multicolumn{11}{c}{\textbf{0.5B+ Models}} \\
\midrule
\graybg{}\textbf{Qwen2.5-Coder-0.5B} & \fst{44.2} & \fst{72.6} & \fst{4.6} & \fst{48.0} & \fst{35.6} & \fst{68.5} & \fst{28.1} & \fst{63.0} \\
\midrule
\multicolumn{11}{c}{\textbf{1B+ Models}} \\
\midrule
DS-Coder-1.3B-Base & 58.7 & 80.4 & 6.2 & 48.8 & 45.8 & 75.0 & 36.9 & 68.1 \\
\graybg{}\textbf{Qwen2.5-Coder-1.5B} & \fst{59.8} & \fst{82.6} & \fst{10.6} & \fst{52.4} & \fst{51.0} & \fst{80.1} & \fst{40.5} & \fst{71.7}
 \\
\midrule
\multicolumn{11}{c}{\textbf{3B+ Models}} \\
\midrule
StarCoder2-3B & 22.3 & 67.4 & 3.1 & 51.6 & 20.6 & 70.1 & 15.3 & 63.0 \\
\graybg{}\textbf{Qwen2.5-Coder-3B} & \fst{64.9} & \fst{85.0} & \fst{12.3} & \fst{55.8} & \fst{54.7} & \fst{81.3} & \fst{44.0} & \fst{74.0} \\
\midrule
\multicolumn{11}{c}{\textbf{6B+ Models}} \\
\midrule
StarCoder2-7B & 19.5 & 67.6 & 4.0 & 53.5 & 19.1 & 72.8 & 14.2 & 64.7 \\
DS-Coder-6.7B-Base & 63.1 & 85.5 & 9.9 & 53.3 & 52.3 & 81.7 & 41.7 & 73.5 \\
DS-Coder-V2-Lite-Base & 66.5 & 85.4 & 10.8 & 53.9 & 53.1 & 81.3 & 43.4 & 73.5 \\
CodeQwen1.5-7B & 59.7 & 81.5 & 4.8 & 44.3 & 46.1 & 77.5 & 36.9 & 67.8 \\
\graybg{}\textbf{Qwen2.5-Coder-7B} & \fst{67.3} & \fst{86.1} & \fst{13.2} & \fst{55.2} & \fst{58.4} & \fst{83.9} & \fst{46.3} & \fst{75.1} \\
\midrule
\multicolumn{11}{c}{\textbf{14B+ Models}} \\
\midrule
StarCoder2-15B & 30.9 & 62.5 & 5.5 & 43.7 & 21.7 & 60.3 & 19.4 & 55.5
 \\
\graybg{}\textbf{Qwen2.5-Coder-14B} & \fst{74.3} & \fst{90.1} & \fst{14.1} & \fst{59.5} & \fst{63.4} & \fst{87.3} & \fst{50.6} & \fst{79.0} \\
\midrule
\multicolumn{11}{c}{\textbf{20B+ Models}} \\
\midrule
 Codestral-22B-v0.1 & 40.9 & 51.7 & 9.9 & 49.2 & 24.8 & 40.8 & 30.0 & 46.6 \\
 DS-Coder-33B-Base & 66.5 & 86.6 & 10.3 & 52.9 & 54.2 & 83.5 & 43.7 & 74.3 \\
\graybg{}\textbf{Qwen2.5-Coder-32B} & \fst{76.1} & \fst{90.5} & \fst{13.6} & \fst{57.5} & \fst{65.1} & \fst{87.6} & \fst{51.6} & \fst{78.5} \\
\bottomrule
\end{tabular}
\caption{Performance of different approaches on the RepoEval Tasks.}
\label{tab:base-repoeval}
\end{table}


SAFIM is a syntax-aware fill-in-the-middle benchmark that emphasizes AST-based code completion, specifically targeting algorithmic blocks, control-flow expressions, and API function calls. The benchmark consists of 17,720 examples from 8,590 code files created after April 2022, deliberately avoiding overlap with mainstream pretraining corpora. For evaluation, we use pass@1 rate as the metric for algorithmic and control-flow tasks, and Exact Match (EM) for API completion tasks.

\subsection{Code Reasoning}
Code is a highly abstract form of logical language, and reasoning based on code helps us determine whether a model truly understands the reasoning flow behind the code. We selected CRUXEval \citep{gu2024cruxeval} as the benchmark, which includes 800 Python functions along with corresponding input-output examples. It consists of two distinct tasks: CRUXEval-I, where the large language model (LLM) must predict the output based on a given input; and CRUXEval-O, where the model must predict the input based on a known output. For both CRUXEval-I and CRUXEval-O, we used a chain-of-thought (CoT) approach, requiring the LLM to output steps sequentially during simulated execution.

As shown in Table \ref{tab:base-cruxeval}, Qwen2.5-Coder delivered highly promising results, achieving a score of 56.5 on CRUXEval-I and 56.0 on CRUXEval-O, thanks to our focus on executable quality during the code cleaning process.

\begin{table}[h]
\centering
\resizebox{0.75\textwidth}{!}{
\begin{tabular}{lr|cc}
\toprule
\multirow{2}{*}{\textbf{Model}} & 
\multirow{2}{*}{\textbf{Size}} & 
\multicolumn{2}{c}{\textbf{CRUXEval}} \\
& &  
\textit{Input-CoT} & \textit{Output-CoT} \\
\midrule
\multicolumn{4}{c}{\textbf{0.5B+ Models}} \\
\midrule
\graybg{}\textbf{Qwen2.5-Coder-0.5B} & 0.5B & \fst{35.2} & \fst{23.0}  \\
\midrule
\multicolumn{4}{c}{\textbf{1B+ Models}} \\
\midrule
DS-Coder-1.3B-Base & 1.3B & 32.1  & 28.2  \\
\graybg{}\textbf{Qwen2.5-Coder-1.5B} & 1.5B & \fst{43.8} & \fst{34.6}  \\
\midrule
\multicolumn{4}{c}{\textbf{3B+ Models}} \\
\midrule
StarCoder2-3B & 3B & 42.1  & 29.2 \\
\graybg{}\textbf{Qwen2.5-Coder-3B} & 3B & \fst{46.5} & \fst{43.8}  \\
\midrule
\multicolumn{4}{c}{\textbf{6B+ Models}} \\
\midrule
StarCoder2-7B & 7B & 39.5 & 35.1 \\
DS-Coder-6.7B-Base & 6.7B & 39.0 & 41.0 \\
DS-Coder-V2-Lite-Base & 2.4/16B & 53.4 & 46.1 \\
CodeQwen1.5-7B & 7B & 44.8 & 40.1 \\
\graybg{}\textbf{Qwen2.5-Coder-7B} & 7B & \fst{56.5}& \fst{56.0} \\
\midrule
\multicolumn{4}{c}{\textbf{14B+ Models}} \\
\midrule
StarCoder2-15B & 15B & 46.1 & 47.6 \\
\graybg{}\textbf{Qwen2.5-Coder-14B} & 14B & \fst{60.6} & \fst{66.4}  \\
\midrule
\multicolumn{4}{c}{\textbf{20B+ Models}} \\
\midrule
 DS-Coder-33B-Base & 33B & 50.6 & 48.8 \\
 DS-Coder-V2-Base & 21/236B & 62.7 & 67.4 \\
\graybg{}\textbf{Qwen2.5-Coder-32B} & 32B & \fst{62.5} & \fst{69.4}  \\
\bottomrule
\end{tabular}
}
\caption{Performance of different models on  CRUXEval with \textit{Input-CoT} and \textit{Output-CoT} settings.}
\label{tab:base-cruxeval}
\end{table}

\subsection{Math Reasoning}
Mathematics and coding have always been closely intertwined. Mathematics forms the foundational discipline for coding, while coding serves as a vital tool in mathematical fields. As such, we expect an open and powerful code model to exhibit strong mathematical capabilities as well. To assess \qwencoder's mathematical performance, we selected five popular benchmarks, including MATH  \citep{hendrycks2021measuringmathematicalproblemsolving}, GSM8K \citep{cobbe2021trainingverifierssolvemath}, MMLU-STEM \citep{hendrycks2021measuringmassivemultitasklanguage} and TheoremQA \citep{chen2023theoremqatheoremdrivenquestionanswering}. Table~\ref{tab:base-matheval} highlights \qwencoder's strengths in mathematics, which likely stem from two key factors: first, the model's strong foundation built on Qwen2.5, and second, the careful mixing of code and mathematical data during training, which has ensured a well-balanced performance across these domains.

\begin{table}[h]
 \centering
 \resizebox{\textwidth}{!}{
 \begin{tabular}{lr|cccc}
 \toprule
 \multirow{2}{*}{\textbf{Model}} & \multirow{2}{*}{\textbf{Size}}  & \textbf{MATH}  & \textbf{GSM8K}  & \textbf{MMLU STEM} & \textbf{TheoremQA} \\
 & & \textit{4-shot} & \textit{4-shot}  & \textit{5-shot}  & \textit{5-shot} \\
  \midrule
 \multicolumn{6}{c}{\textbf{0.5B+ Models}} \\
 \midrule
 \graybg{}\textbf{Qwen2.5-Coder-0.5B}  & 0.5B  & \fst{15.4} & \fst{34.5} & \fst{34.4} & \fst{14.3} \\
 \midrule
 \multicolumn{6}{c}{\textbf{1B+ Models}} \\
 \midrule
 DS-Coder-1.3B-Base  & 1.3B  &  4.6 & 4.4 & 24.5 & 8.9 \\
\graybg{}\textbf{Qwen2.5-Coder-1.5B}  & 1.5B  &  \fst{30.9} & \fst{65.8} & \fst{49.0} & \fst{21.4} \\
 \midrule
 \multicolumn{6}{c}{\textbf{3B+ Models}} \\
 \midrule
  StarCoder2-3B  & 3B  & 10.8 & 21.6 & 34.9 & 12.1 \\
 \graybg{}\textbf{Qwen2.5-Coder-3B}  & 3B  & \fst{40.0} & \fst{75.7} & \fst{56.0} & \fst{29.5} \\
 \midrule
 \multicolumn{6}{c}{\textbf{6B+ Models}} \\
 \midrule
 StarCoder2-7B  & 7B  &  14.6 & 32.7 & 39.8 & 16.0 \\
 DS-Coder-6.7B-Base  & 6.7B  &  10.3 & 21.3 & 34.2 & 13.6 \\
 DS-Coder-V2-Lite-Base  & 2.4/16B  &  39.0 & 67.1 & 58.5 & 29.3 \\
 CodeQwen1.5-7B  & 7B  &  10.6 & 37.7 & 39.6 & 15.8 \\
 \graybg{}\textbf{Qwen2.5-Coder-7B}  & 7B  &  \fst{46.6} & \fst{83.9} & \fst{67.6} & \fst{34.0} \\
  \midrule
 \multicolumn{6}{c}{\textbf{14B+ Models}} \\
   \midrule
  StarCoder2-15B  & 15B  &  23.7 & 57.7 & 49.2 & 20.5 \\
  \graybg{}\textbf{Qwen2.5-Coder-14B}  & 14B  & \fst{52.8} & \fst{88.7} & \fst{73.9} & \fst{39.6} \\
 \midrule
 \multicolumn{6}{c}{\textbf{20B+ Models}} \\
 \midrule
  DS-Coder-33B-Base  & 33B  & 14.4 & 35.4 & 39.5 & 17.5 \\
  DS-Coder-V2-Base  & 21/236B  & 50.6 & 85.8 & \fst{76.0} & 39.4\\
 \graybg{}\textbf{Qwen2.5-Coder-32B}  & 32B  &  \fst{57.2} & \fst{91.1} & 75.1 & \fst{43.1} \\
 \bottomrule
 \end{tabular}
 }
 \caption{Performance of various models on four math benchmarks, named MATH, GSM8K, MMLU STEM and TheoremQA respectively. }
 \label{tab:base-matheval}
\end{table}

\begin{table}[h]
\centering
\begin{tabular}{lr|ccc}
\toprule
\multirow{2}{*}{\textbf{Model}} & 
\multirow{2}{*}{\textbf{Size}} & 
\multicolumn{3}{c}{\textbf{MMLU}} \\
& & \textbf{Base}& \textbf{Pro} & \textbf{Redux}  \\
\midrule
\multicolumn{5}{c}{\textbf{0.5B+ Models}} \\
\midrule
\graybg \textbf{Qwen2.5-Coder-0.5B} & 0.5B & \fst{42.0} & \fst{13.3} & \fst{40.6} \\
\midrule
\multicolumn{5}{c}{\textbf{1B+ Models}} \\
\midrule
DS-Coder-1.3B-Base & 1.3B & 25.8 & 11.4 & 24.5 \\
\graybg \textbf{Qwen2.5-Coder-1.5B} & 1.5B &\fst{53.6} & \fst{23.1} & \fst{50.9} \\
\midrule
\multicolumn{5}{c}{\textbf{3B+ Models}} \\
\midrule
StarCoder2-3B & 3B & 36.6 & 15.5 & 37.0 \\
\graybg \textbf{Qwen2.5-Coder-3B} & 3B & \fst{61.2} & \fst{32.0} & \fst{59.5}  \\
\midrule
\multicolumn{5}{c}{\textbf{6B+ Models}} \\
\midrule
StarCoder2-7B & 7B & 38.8 & 17.2 & 38.6 \\
DS-Coder-6.7B-Base & 6.7B & 36.4 & 16.7 & 36.5 \\
DS-Coder-V2-Lite-Base & 2.4/16B & 60.5 & 33.4 & 58.3 \\
CodeQwen1.5-7B & 7B & 40.5 & 17.2 & 41.2 \\
\graybg \textbf{Qwen2.5-Coder-7B} & 7B & \fst{68.0} & \fst{40.1} & \fst{66.6} \\
\midrule
\multicolumn{5}{c}{\textbf{14B+ Models}} \\
\midrule
StarCoder2-15B & 15B & 64.1 & 24.3 & 48.8 \\
\graybg \textbf{Qwen2.5-Coder-14B} & 14B & \fst{75.2} & \fst{49.3} & \fst{72.4}  \\
\midrule
\multicolumn{5}{c}{\textbf{20B+ Models}} \\
\midrule
 DS-Coder-33B-Base & 33B & 39.4 & 18.4 & 38.7 \\
\graybg \textbf{Qwen2.5-Coder-32B} & 32B &  \fst{79.1} & \fst{50.4} & \fst{77.5} \\
\bottomrule
\end{tabular}
\caption{MMLU results of different models, a general benchmark for common knowledge.}
\label{tab:base-general-mmlu}
\end{table}

\begin{table}[h]
\centering
\resizebox{\textwidth}{!}{
\begin{tabular}{lr|cccc}
\toprule
\textbf{Model} & \textbf{Size} & 
\textbf{ARC-Challenge} & \textbf{TruthfulQA} & \textbf{WinoGrande} & \textbf{HellaSwag} \\
\midrule
\multicolumn{6}{c}{\textbf{0.5B+ Models}} \\
\midrule
\graybg \textbf{Qwen2.5-Coder-0.5B} & 0.5B & \fst{34.4} & \fst{42.7} & \fst{54.8} & \fst{48.4} \\
\midrule
\multicolumn{6}{c}{\textbf{1B+ Models}} \\
\midrule
DS-Coder-1.3B-Base & 1.3B & 25.4  & 42.7  & 53.3  & 39.5 \\
\graybg \textbf{Qwen2.5-Coder-1.5B} & 1.5B & \fst{45.2} & \fst{44.0} & \fst{60.7} & \fst{61.8} \\
\midrule
\multicolumn{6}{c}{\textbf{3B+ Models}} \\
\midrule
StarCoder2-3B & 3B & 34.2  & 40.5  & 57.1  & 48.1 \\
\graybg \textbf{Qwen2.5-Coder-3B} & 3B & \fst{52.9} & \fst{49.2} & \fst{67.4} & \fst{70.9} \\
\midrule
\multicolumn{6}{c}{\textbf{6B+ Models}} \\
\midrule
StarCoder2-7B & 7B & 38.7  & 42.0  & 57.1  & 52.4 \\
DS-Coder-6.7B-Base & 6.7B & 36.4  & 40.2  & 57.6  & 53.8 \\
DS-Coder-V2-Lite-Base & 2.4/16B & 57.3  & 38.8  & \fst{72.9} & 76.1 \\
CodeQwen1.5-7B & 7B & 35.7  & 42.2  & 59.8  & 56.0 \\
\graybg \textbf{Qwen2.5-Coder-7B} & 7B & \fst{60.9} & \fst{50.6} & \fst{72.9} & \fst{76.8} \\
\midrule
\multicolumn{6}{c}{\textbf{14B+ Models}} \\
\midrule
StarCoder2-15B & 15B & 47.2  & 37.9  & 64.3  & 64.1 \\
\graybg \textbf{Qwen2.5-Coder-14B} & 14B & \fst{66.0} & \fst{55.2} & \fst{76.8} & \fst{80.2} \\
\midrule
\multicolumn{6}{c}{\textbf{20B+ Models}} \\
\midrule
 DS-Coder-33B-Base &  33B &  42.2  &  40.0  &  62.0  & 60.2 \\
 DS-Coder-V2-Base & 21/236B  &  64.3  &  41.4  & \fst{83.7} & \fst{86.0} \\
\graybg \textbf{Qwen2.5-Coder-32B} & 32B & \fst{70.5} & \fst{54.2} & 80.8 & 83.0 \\
\bottomrule
\end{tabular}
}
\caption{General performance of different models on four popular general benchmarks, ARC-Challenge, TruthfulQA, WinoGrande and HellaSwag.}
\label{tab:base-general-others}
\end{table}

\subsection{General Natural Language}
In addition to mathematical ability, we aim to retain as much of the base model’s general-purpose capabilities as possible, such as general knowledge. To evaluate general natural language understanding, we selected MMLU  \citep{hendrycks2021measuringmathematicalproblemsolving} and its variant MMLU-Redux \citep{gema2024mmlu}, along with four other benchmarks: ARC-Challenge \citep{clark2018thinksolvedquestionanswering}, TruthfulQA \citep{lin2022truthfulqameasuringmodelsmimic}, WinoGrande \citep{sakaguchi2019winograndeadversarialwinogradschema}, and HellaSwag \citep{zellers2019hellaswagmachinereallyfinish}. Similar to the results in mathematics, Table \ref{tab:base-general-others} highlights \qwencoder's advantage in general natural language capabilities compared to other coders, further validating the effectiveness of \qwencoder data mixing strategy.

\subsection{Long-Context Evaluation}
Long context capability is crucial for code LLMs, serving as the core skill for understanding repository-level code and becoming a code agent. However, most of the current code models still have very limited support for length, which hinders their potential for practical application. \qwencoder aims to further advance the progress of open-source code models in long context modeling. To achieve this, we have collected and constructed long sequence code data at the repository level for pre-training. Through careful data proportioning and organization, we have enabled it to support input lengths of up to 128K tokens.

\paragraph{Needle in the Code}
We created a simple but basic synthetic task called \textit{Needle in the Code}, inspired by popular long-context evaluations in the text domain. In this task, we inserted a very simple custom function at various positions within a code repo (we chose Megatron \footnote{\url{https://github.com/NVIDIA/Megatron-LM}} to honor its contributions to open-source LLMs!) and tested whether the model could replicate this function at the end of the codebase. The figure below shows that \qwencoder is capable of successfully completing this task within a 128k length range.

\begin{figure}[htbp]
    \centering
    \includegraphics[width=1.0\columnwidth]{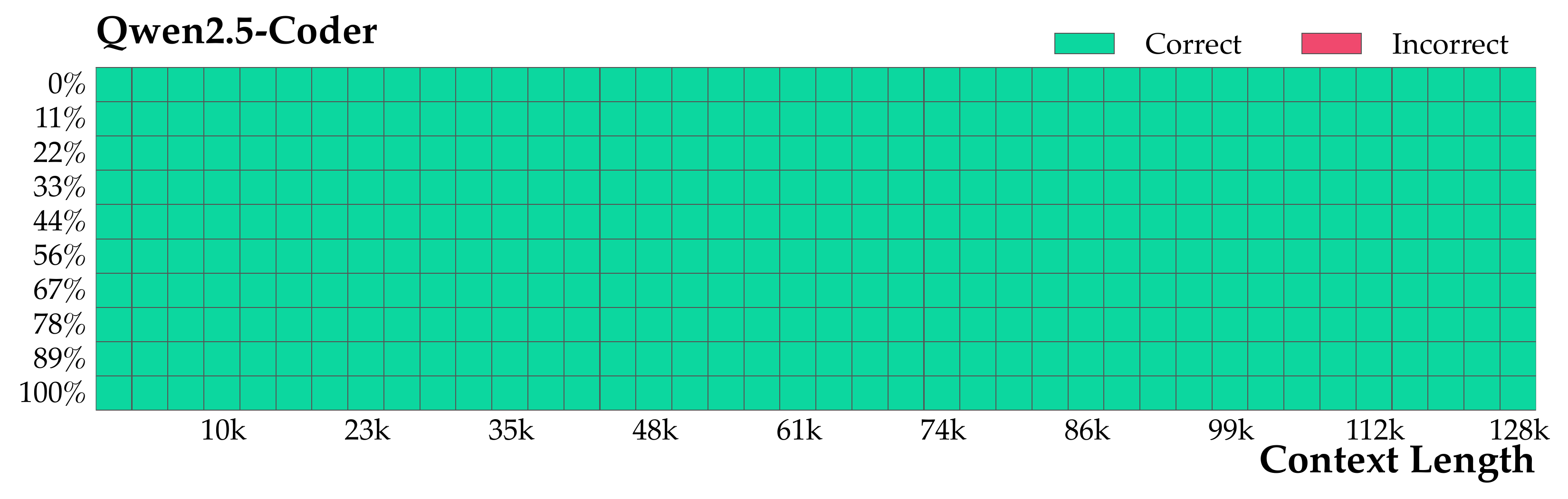}
    \vspace{-1mm}
    \caption{The long context ability of \qwencoder, evaluated by Needle in the Code.}
    \label{fig:needle}
\end{figure}

\section{Evaluation on Instruct Models}

For the evaluation of the instruct models, we rigorously assessed six core areas: \textit{code generation}, \textit{code reasoning}, \textit{code editing}, \textit{text-to-sql}, \textit{mathematical reasoning} and \textit{general natural language understanding}. The evaluation was structured to ensure a fair and thorough comparison across models. All evaluation code is publicly accessible for reproducibility\footnote{\url{https://github.com/QwenLM/Qwen2.5-Coder}}. To ensure a broad comparison, we included some of the most popular and widely-used open-source instruction-tuned models, notably versions from the DeepSeek-Coder series and Codestral models. Below is a list of all artifacts referenced in this section.

\begin{table}[htbp]
 \small
 \centering
 \begin{tabular}{l|l}
 \toprule
 \textbf{Artifact} & \textbf{Public link} \\
 \midrule
 Qwen2.5-Coder-0.5B-Instruct & \url{https://hf.co/Qwen/Qwen2.5-Coder-0.5B-Instruct} \\
 Qwen2.5-Coder-1.5B-Instruct & \url{https://hf.co/Qwen/Qwen2.5-Coder-1.5B-Instruct} \\
 Qwen2.5-Coder-3B-Instruct & \url{https://hf.co/Qwen/Qwen2.5-Coder-3B-Instruct} \\
 Qwen2.5-Coder-7B-Instruct & \url{https://hf.co/Qwen/Qwen2.5-Coder-7B-Instruct} \\
 Qwen2.5-Coder-14B-Instruct & \url{https://hf.co/Qwen/Qwen2.5-Coder-14B-Instruct} \\
 Qwen2.5-Coder-32B-Instruct & \url{https://hf.co/Qwen/Qwen2.5-Coder-32B-Instruct} \\
 \midrule
 CodeQwen1.5-7B-Chat & \url{https://hf.co/Qwen/CodeQwen1.5-7B-Chat} \\
 CodeLlama-7B-Instruct & \url{https://hf.co/meta-llama/CodeLlama-7b-Instruct-hf} \\
 CodeLlama-13B-Instruct & \url{https://hf.co/meta-llama/CodeLlama-13b-Instruct-hf} \\
 CodeLlama-34B-Instruct & \url{https://hf.co/meta-llama/CodeLlama-34b-Instruct-hf} \\
 CodeLlama-70B-Instruct & \url{https://hf.co/meta-llama/CodeLlama-70b-Instruct-hf} \\
 DS-Coder-1.3B-instruct & \url{https://hf.co/deepseek-ai/deepseek-coder-1.3b-instruct} \\
 DS-Coder-6.7B-instruct & \url{https://hf.co/deepseek-ai/deepseek-coder-6.7b-instruct} \\
 DS-Coder-33B-instruct & \url{https://hf.co/deepseek-ai/deepseek-coder-33b-instruct} \\
 DS-Coder-V2-Lite-Instruct & \url{https://hf.co/deepseek-ai/DeepSeek-Coder-V2-Lite-Instruct} \\
 DS-Coder-V2-Instruct & \url{https://hf.co/deepseek-ai/DeepSeek-Coder-V2-Instruct} \\
 Starcoder2-15B-Instruct-v0.1 & \url{https://hf.co/bigcode/starcoder2-15b-instruct-v0.1} \\
 CodeStral-22B-v0.1 & \url{https://hf.co/mistralai/Codestral-22B-v0.1} \\
 Yi-Coder-1.5B-Chat & \url{https://hf.co/01-ai/Yi-Coder-1.5B-Chat} \\
 Yi-Coder-9B-Chat & \url{https://hf.co/01-ai/Yi-Coder-9B-Chat} \\
 \bottomrule
 \end{tabular}
 \caption{All artifacts released and used in this section.}
 \label{tab:artifacts_instruct}
\end{table}

\subsection{Code Generation}
Building on the performance improvements of the Qwen2.5-Coder series base models, our Qwen2.5-Coder series instruct models similarly demonstrated outstanding performance in code generation tasks.

\paragraph{HumanEval and MBPP} 

We also assessed the code generation capabilities of the Qwen2.5-Coder series instruction models using the EvalPlus \citep{evalplus} dataset. As shown by the results in Table \ref{tab:instruct-codegen}, our Qwen2.5-Coder-7B-Instruct model demonstrated exceptional accuracy, significantly outperforming other models with a comparable parameter count. Remarkably, it even surpassed larger models with over 20 billion parameters, such as CodeStral-22B and DS-Coder-33B-Instruct. Furthermore, our Qwen2.5-Coder-32B-Instruct model achieved the highest performance on EvalPlus, even outperforming DS-Coder-V2-Instruct, making it the most powerful open-source code model to date.

\begin{table}[h]
    \centering
    \resizebox{\textwidth}{!}{
        \begin{tabular}{lr|cc|cc|cc|c}
            \toprule
            \multirow{2}{*}{\textbf{Model}}              &
            \multirow{2}{*}{\textbf{Size}}               &
            \multicolumn{2}{c|}{\textbf{HumanEval}}      &
            \multicolumn{2}{c|}{\textbf{MBPP}}           &
            \multicolumn{2}{c|}{\textbf{BigCodeBench}}   &
            \textbf{LiveCodeBench}                                                                                                                                                                                    \\
                                                         &                &
            \textit{HE}                                  & \textit{HE+}   &
            \textit{MBPP}                                & \textit{MBPP+} &
            \textit{Full}                                & \textit{Hard}  &
            \textit{Pass@1}
            \\
            \midrule
            \multicolumn{9}{c}{\textbf{0.5B+ Models}}                                                                                                                                                                 \\
            \midrule
            \graybg \textbf{Qwen2.5-Coder-0.5B-Instruct} & 0.5B           & \fst{61.6}        & \fst{57.3}        & \fst{52.4}        & \fst{43.7}        & \fst{11.1}        & \fst{1.4}         & \fst{2.0}         \\
            \midrule
            \multicolumn{9}{c}{\textbf{1B+ Models}}                                                                                                                                                                   \\
            \midrule
            DS-Coder-1.3B-Instruct                       & 1.3B           & 65.9              & 60.4              & 65.3              & 54.8              & 22.8              & 3.4               & 5.1               \\
            Yi-Coder-1.5B-Chat                           & 1.5B           & 69.5              & 64.0              & 65.9              & 57.7              & 23.8              & \fst{11.5}        & 4.8               \\
            \graybg \textbf{Qwen2.5-Coder-1.5B-Instruct} & 1.5B           & \fst{70.7}        & \fst{66.5}        & \fst{69.2}        & \fst{59.4}        & \fst{32.5}        & 6.8               & \fst{6.1}         \\
            \midrule
            \multicolumn{9}{c}{\textbf{3B+ Models}}                                                                                                                                                                   \\
            \midrule
            \graybg \textbf{Qwen2.5-Coder-3B-Instruct}   & 3B             & \fst{84.1}        & \fst{80.5}        & \fst{73.6}        & \fst{62.4}        & \fst{35.8}        & \fst{14.2}        & \fst{10.8}        \\
            \midrule
            \multicolumn{9}{c}{\textbf{6B+ Models}}                                                                                                                                                                   \\ \midrule
            CodeLlama-7B-Instruct                        & 7B             & 40.9              & 33.5              & 54.0              & 44.4              & 21.9              & 3.4               & 7.1               \\
            DS-Coder-6.7B-Instruct                       & 6.7B           & 74.4              & 71.3              & 74.9              & 65.6              & 35.5              & 10.1              & 15.5              \\
            CodeQwen1.5-7B-Chat                          & 7B             & 83.5              & 78.7              & 77.7              & 67.2              & 39.6              & 18.9              & 7.9               \\
            Yi-Coder-9B-Chat                             & 9B             & 82.3              & 74.4              & 82.0              & 69.0              & 38.1              & 11.5              & 17.2              \\
            DS-Coder-V2-Lite-Instruct                    & 2.4/16B        & 81.1              & 75.6              & 82.8              & 70.4              & 36.8              & 16.2              & 16.3              \\
            \graybg \textbf{Qwen2.5-Coder-7B-Instruct}   & 7B             & \fst{88.4}        & \fst{84.1}        & \fst{83.5}        & \fst{71.7}        & \fst{41.0}        & \fst{18.2}        & \fst{18.2}        \\
            \midrule
            \multicolumn{9}{c}{\textbf{13B+ Models}}                                                                                                                                                                  \\
            \midrule
            CodeLlama-13B-Instruct                       & 13B            & 40.2              & 32.3              & 60.3              & 51.1              & 28.5              & 9.5               & 6.1               \\
            Starcoder2-15B-Instruct-v0.1                 & 15B            & 67.7              & 60.4              & 78.0              & 65.1              & 37.2              & 11.5              & 12.1              \\
            \graybg \textbf{Qwen2.5-Coder-14B-Instruct}  & 14B            & \fst{89.6}        & \fst{87.2}        & \fst{86.2}        & \fst{72.8}        & \fst{48.4}        & \fst{22.2}         & \fst{23.4}        \\
            \midrule
            \multicolumn{9}{c}{\textbf{20B+ Models}}                                                                                                                                                                  \\ \midrule
            CodeLlama-34B-Instruct                       & 34B            & 48.2              & 40.2              & 61.1              & 50.5              & 29.0              & 8.8               & 8.4               \\
            CodeStral-22B-v0.1                           & 22B            & 81.1              & 73.2              & 78.2              & 62.2              & 41.8              & 16.9              & 22.6              \\
            DS-Coder-33B-Instruct                        & 33B            & 81.1              & 75.0              & 80.4              & 70.1              & 42.0              & 17.6              & 21.3              \\
            CodeLlama-70B-Instruct                       & 70B            & 72.0              & 65.9              & 77.8              & 64.6              & 40.7              & 11.5              & 3.3               \\
            DS-Coder-V2-Instruct                         & 21/236B        & 85.4              & 82.3              & 89.4              & \fst{75.1}        & 48.2              & 24.3              & 27.9              \\
            \graybg \textbf{Qwen2.5-Coder-32B-Instruct}  & 32B            & \fst{92.7}        & \fst{87.2}        & \fst{90.2}        & \fst{75.1}        & \fst{49.6}        & \fst{27.0}        & \fst{31.4}        \\
            \midrule
            \multicolumn{9}{c}{\grayt \textbf{Closed-APIs}}                                                                                                                                                                  \\ \midrule
            \grayt Claude-3.5-Sonnet-20240620            & \grayt -       & \grayt 89.0       & \grayt 81.1       & \grayt 87.6       & \grayt 72.0       & \grayt 45.3       & \grayt 25.7       & \grayt 32.1       \\
            \grayt Claude-3.5-Sonnet-20241022            & \grayt -       & \grayt 92.1       & \grayt 86.0       & \grayt 91.0       & \grayt 74.6       & \grayt 45.3       & \grayt 23.6       & \grayt 31.6       \\
            \grayt GPT-4o-mini-2024-07-18                & \grayt -       & \grayt 87.8       & \grayt 84.8       & \grayt 86.0       & \grayt 72.2       & \grayt 46.9       & \grayt 23.6       & \grayt 28.3       \\
            \grayt GPT-4o-2024-08-06                     & \grayt -       & \grayt 92.1       & \grayt 86.0       & \grayt 86.8       & \grayt 72.5       & \grayt \fst{50.1} & \grayt 25.0       & \grayt 34.6       \\
            \grayt o1-mini                               & \grayt -       & \grayt \fst{97.6} & \grayt \fst{90.2} & \grayt \fst{93.9} & \grayt \fst{78.3} & \grayt 46.3       & \grayt 23.0       & \grayt \fst{60.0} \\
            \grayt o1-preview                            & \grayt -       & \grayt 95.1       & \grayt 88.4       & \grayt 93.4       & \grayt 77.8       & \grayt 49.3       & \grayt \fst{27.7} & \grayt 43.1       \\
            \bottomrule
        \end{tabular}
    }
    \caption{The performance of different instruct
        models on code generation by HumanEval, MBPP, bigcodebench and livecodebench. For bigcodebench here, we report ``instruct'' tasks score.}
    \label{tab:instruct-codegen}
\end{table}

\begin{table}[h]
    \centering
    \resizebox{\textwidth}{!}{
        \begin{tabular}{lr|cccccccc|c}
            \toprule
            \textbf{Model}                               & \textbf{Size} &
            Python                                       & Java          & C++               & C\#               & TS                & JS                & PHP               & Bash              & \textbf{Average}                                          \\
            \midrule
            \multicolumn{11}{c}{\textbf{0.5B+ Models}}                                                                                                                                                                                                       \\ \midrule
            \graybg \textbf{Qwen2.5-Coder-0.5B-Instruct} & 0.5B          & \fst{62.8}        & \fst{46.2}        & \fst{43.5}        & \fst{62.7}        & \fst{50.3}        & \fst{50.3}        & \fst{52.8}        & \fst{27.8}        & \fst{49.6}        \\
            \midrule
            \multicolumn{11}{c}{\textbf{1B+ Models}}                                                                                                                                                                                                         \\ \midrule
            DS-Coder-1.3B-Instruct                       & 1.3B          & 65.2              & 51.9              & 45.3              & 55.1              & 59.7              & 52.2              & 45.3              & 12.7              & 48.4              \\
            Yi-Coder-1.5B-Chat                           & 1.5B          & 67.7              & 51.9              & 49.1              & 57.6              & 57.9              & 59.6              & 52.2              & 19.0              & 51.9              \\
            \graybg \textbf{Qwen2.5-Coder-1.5B-Instruct} & 1.5B          & \fst{71.2}        & \fst{55.7}        & \fst{50.9}        & \fst{64.6}        & \fst{61.0}        & \fst{62.1}        & \fst{59.0}        & \fst{29.1}        & \fst{56.7}        \\
            \midrule
            \multicolumn{11}{c}{\textbf{3B+ Models}}                                                                                                                                                                                                         \\ \midrule
            \graybg \textbf{Qwen2.5-Coder-3B-Instruct}   & 3B            & \fst{83.5}        & \fst{74.7}        & \fst{68.3}        & \fst{78.5}        & \fst{79.9}        & \fst{75.2}        & \fst{73.3}        & \fst{43.0}        & \fst{72.1}        \\
            \midrule
            \multicolumn{11}{c}{\textbf{6B+ Models}}                                                                                                                                                                                                         \\ \midrule
            CodeLlama-7B-Instruct                        & 7B            & 34.8              & 30.4              & 31.1              & 21.6              & 32.7              & -                 & 28.6              & 10.1              & -                 \\
            DS-Coder-6.7B-Instruct                       & 6.7B          & 78.6              & 68.4              & 63.4              & 72.8              & 67.2              & 72.7              & 68.9              & 36.7              & 66.1              \\
            CodeQwen1.5-7B-Chat                          & 7B            & 84.1              & 73.4              & 74.5              & 77.8              & 71.7              & 75.2              & 70.8              & 39.2              & 70.8              \\
            Yi-Coder-9B-Chat                             & 9B            & 85.4              & 76.0              & 67.7              & 76.6              & 72.3              & 78.9              & 72.1              & 45.6              & 71.8              \\
            DS-Coder-V2-Lite-Instruct                    & 2.4/16B       & 81.1              & 76.6              & 75.8              & 76.6              & 80.5              & 77.6              & 74.5              & 43.0              & 73.2              \\
            \graybg \textbf{Qwen2.5-Coder-7B-Instruct}   & 7B            & \fst{87.8}        & \fst{76.5}        & \fst{75.6}        & \fst{80.3}        & \fst{81.8}        & \fst{83.2}        & \fst{78.3}        & \fst{48.7}        & \fst{76.5}        \\
            \midrule
            \multicolumn{11}{c}{\textbf{13B+ Models}}                                                                                                                                                                                                        \\ \midrule
            CodeLlama-13B-Instruct                       & 13B           & 42.7              & 40.5              & 42.2              & 24.0              & 39.0              & -                 & 32.3              & 13.9              & -                 \\
            Starcoder2-15B-Instruct-v0.1                 & 15B           & 68.9              & 53.8              & 50.9              & 62.7              & 57.9              & 59.6              & 53.4              & 24.7              & 54.0              \\
            \graybg \textbf{Qwen2.5-Coder-14B-Instruct}  & 14B           & \fst{89.0}        & \fst{79.7}        & \fst{85.1}        & \fst{84.2}        & \fst{86.8}        & \fst{84.5}        & \fst{80.1}        & \fst{47.5}        & \fst{79.6}        \\
            \midrule
            \multicolumn{11}{c}{\textbf{20B+ Models}}                                                                                                                                                                                                        \\ \midrule
            CodeLlama-34B-Instruct                       & 34B           & 41.5              & 43.7              & 45.3              & 31.0              & 40.3              & -                 & 36.6              & 19.6              & -                 \\
            CodeStral-22B-v0.1                           & 22B           & 81.1              & 63.3              & 65.2              & 43.7              & 68.6              & -                 & 68.9              & 42.4              & -                  \\
            DS-Coder-33B-Instruct                        & 33B           & 79.3              & 73.4              & 68.9              & 74.1              & 67.9              & 73.9              & 72.7              & 43.0              & 69.2              \\
            CodeLlama-70B-Instruct                       & 70B           & 67.8              & 58.2              & 53.4              & 36.7              & 39.0              & -                 & 58.4              & 29.7              & -                 \\
            DS-Coder-V2-Instruct                         & 21/236B       & 90.2              & \fst{82.3}        & \fst{84.8}        & 82.3              & 83.0              & 84.5              & \fst{79.5}        & \fst{52.5}        & \fst{79.9}        \\
            \graybg \textbf{Qwen2.5-Coder-32B-Instruct}  & 32B           & \fst{92.7}        & 80.4              & 79.5              & \fst{82.9}        & \fst{86.8}        & \fst{85.7}        & 78.9              & 48.1              & 79.4              \\
            \midrule
            \multicolumn{11}{c}{\grayt \textbf{Closed-APIs}}                                                                                                                                                                                                        \\ \midrule
            \grayt Claude-3.5-Sonnet-20240620            & \grayt -      & \grayt 89.6       & \grayt 86.1       & \grayt 82.6       & \grayt 85.4       & \grayt 84.3       & \grayt 84.5       & \grayt 80.7       & \grayt 48.1       & \grayt 80.2       \\
            \grayt Claude-3.5-Sonnet-20241022            & \grayt -      & \grayt 93.9       & \grayt 86.7       & \grayt 88.2       & \grayt \fst{87.3} & \grayt 88.1       & \grayt 91.3       & \grayt 82.6       & \grayt 52.5       & \grayt 83.8       \\
            \grayt GPT-4o-mini-2024-07-18                & \grayt -      & \grayt 87.2       & \grayt 75.9       & \grayt 77.6       & \grayt 79.7       & \grayt 79.2       & \grayt 81.4       & \grayt 75.2       & \grayt 43.7       & \grayt 75.0       \\
            \grayt GPT-4o-2024-08-06                     & \grayt -      & \grayt 90.9       & \grayt 83.5       & \grayt 76.4       & \grayt 81.0       & \grayt 83.6       & \grayt 90.1       & \grayt 78.9       & \grayt 48.1       & \grayt 79.1       \\
            \grayt o1-mini                               & \grayt -      & \grayt 95.7       & \grayt \fst{90.5} & \grayt \fst{93.8} & \grayt 77.2       & \grayt \fst{91.2} & \grayt 92.5       & \grayt 84.5       & \grayt \fst{55.1} & \grayt 85.1       \\
            \grayt o1-preview                            & \grayt -      & \grayt \fst{96.3} & \grayt 88.0       & \grayt 91.9       & \grayt 84.2       & \grayt 90.6       & \grayt \fst{93.8} & \grayt \fst{90.1} & \grayt 47.5       & \grayt \fst{85.3} \\
            \bottomrule
        \end{tabular}
    }
    \caption{The performance of different models on instruct format MultiPL-E.}
    \label{tab:instruct-multiple}
\end{table}

\paragraph{BigCodeBench-Instruct}

The \emph{instruct} split provided by BigCodeBench \citep{zhuo2024bigcodebenchbenchmarkingcodegeneration} is designed to evaluate the code generation capabilities of instruction-based models. We evaluated the Qwen2.5-Coder series instruct models on the BigCodeBench-Instruct dataset. As indicated in Table \ref{tab:instruct-codegen}, the Qwen2.5-Coder-7B-Instruct model outperformed other instruct models with comparable parameter sizes, achieving notably high accuracy scores on both the full and hard subsets, reaching 41.0\% on the full subset and 18.2\% on the hard subset. This highlights the robust code generation capabilities of the Qwen2.5-Coder instruct models. Furthermore, the Qwen2.5-Coder-32B-Instruct achieved accuracy rates of 49.6\% on the complete split and 27.0\% on the hard split, establishing it as the best-performing open-source code generation model and surpassing several closed-source APIs.

\paragraph{LiveCodeBench}

LiveCodeBench \citep{jain2024livecodebench} is a comprehensive and contamination-free benchmark designed to evaluate the coding capabilities of LLMs. It continuously gathers new problems from leading competitive programming platforms like LeetCode\footnote{\url{https://leetcode.com}}, AtCoder\footnote{\url{https://atcoder.jp}}, and CodeForces\footnote{\url{https://codeforces.com}}, ensuring an up-to-date and diverse set of challenges. Currently, it hosts over 600 high-quality coding problems published between May 2023 and September 2024.

To further demonstrate our model’s effectiveness on real-world competitive programming tasks, we evaluated the Qwen-2.5-Coder series instruct models on the LiveCodeBench (2407-2409) dataset. As shown in Table \ref{tab:instruct-codegen}, the Qwen-2.5-Coder-7B-Instruct model achieved an impressive Pass@1 accuracy of 37.6\%, significantly outperforming other models with similar parameter counts. Notably, it also outperformed larger models, such as CodeStral-22B-v0.1 and DS-Coder-33B-Instruct. Additionally, our Qwen-2.5-Coder-32B-Instruct model achieved an accuracy of 31.4\%, surpassing all open-source code generation models and reaching a level comparable to many closed-source APIs.


\paragraph{Multi-Programming Language}
The Qwen2.5-Coder series instruct models have inherited the high performance of the base model on the Multi-Programming Language. To further evaluate their capabilities, we tested the instruct models on two specific benchmarks: MultiPL-E \citep{cassano2022multiplescalableextensibleapproach} and McEval \citep{chai2024mceval}.

\paragraph{MultiPL-E}

As shown by the evaluation results in Table \ref{tab:instruct-multiple}, Qwen2.5-Coder-7B-Instruct consistently outperforms other models with similar parameter counts, such as DS-Coder-V2-Lite-Instruct, in code generation tasks across eight programming languages. Both Qwen2.5-Coder-7B-Instruct and Qwen2.5-Coder-14B-Instruct even surpass larger models, like CodeStral-22B and DS-Coder-33B-Instruct (which have over 20 billion parameters), underscoring their strong code generation capabilities across multiple languages. Our Qwen2.5-Coder-32B-Instruct model achieves comparable performance to the DS-Coder-V2-Instruct model with only 32 billion parameters, bringing it very close to the performance of several closed-source APIs.

\begin{figure}[h]
    \centering
    \includegraphics[width=0.9\columnwidth]{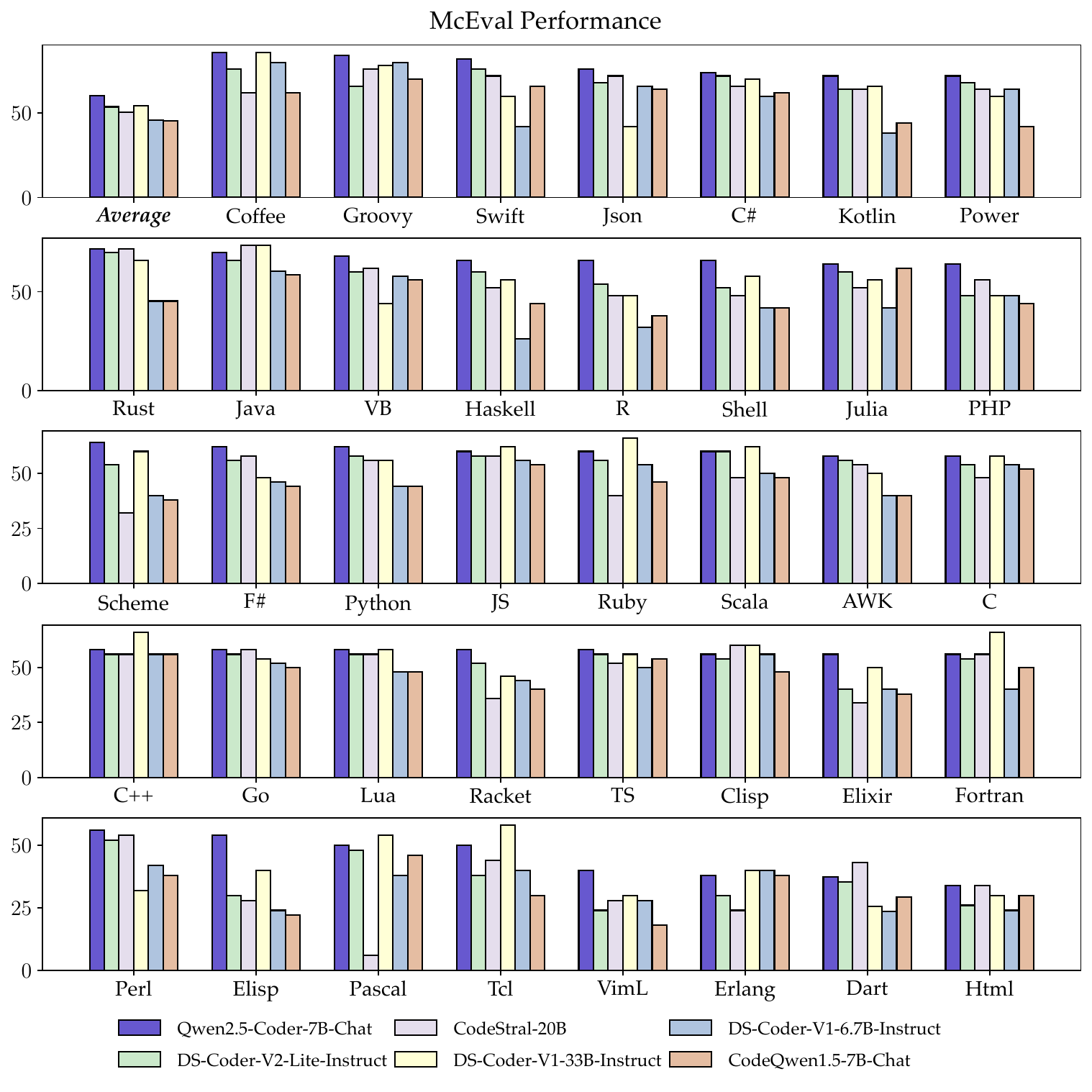}
    \vspace{-1mm}
    \caption{The McEval Performance of Qwen2.5-Coder-32B-Instruct compared with popular open-source large code models with similar size.}
    \label{fig:instruct-mceval}
\end{figure}

\begin{figure}[h]
    \centering
    \includegraphics[width=0.9\columnwidth]{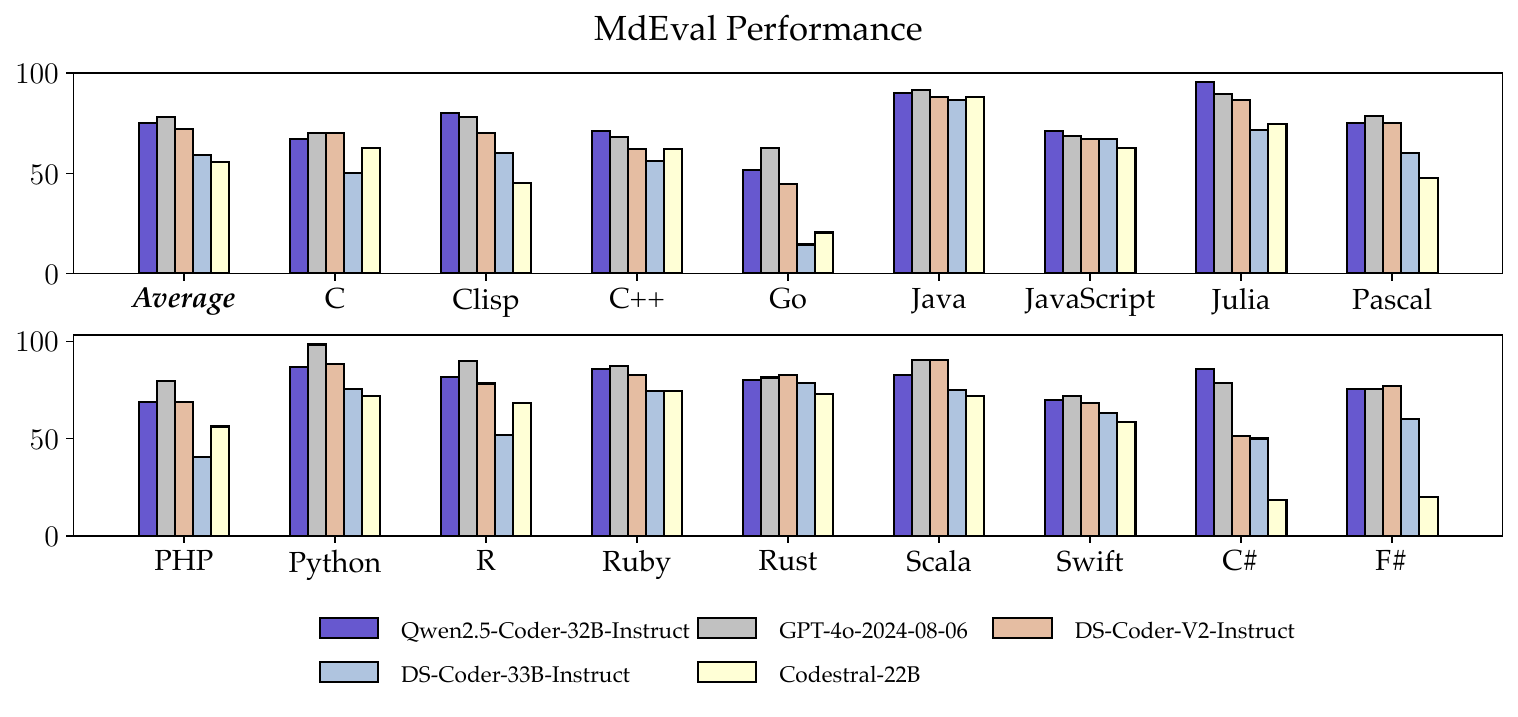}
    \caption{The MdEval Performance of Qwen2.5-Coder-32B-Instruct compared with popular open-source large code models with similar size.}
    \label{fig:instruct-mdeval}
\end{figure}

\paragraph{McEval}
To comprehensively assess the code generation capabilities of the Qwen2.5-Coder series models across a broader range of programming languages, we evaluated them on the McEval benchmark \citep{chai2024mceval}, which spans 40 programming languages and includes 16,000 test cases.
As shown in Figure \ref{fig:instruct-mceval}, the Qwen2.5-Coder-32B-Instruct model excels when compared to other open-source models on the McEval benchmark, particularly across a wide range of programming languages.

\begin{table}[h]
    \centering
    \resizebox{0.72\textwidth}{!}{
        \begin{tabular}{lr|cc}
            \toprule
            \multirow{2}{*}{\textbf{Model}}              &
            \multirow{2}{*}{\textbf{Size}}               &
            \multicolumn{2}{c}{\textbf{CRUXEval}}                                                          \\
                                                         &                     &
            \textit{Input-CoT}                           & \textit{Output-CoT}                             \\
            \midrule
            \multicolumn{4}{c}{\textbf{0.5B+ Models}}                                                      \\ \midrule
            \graybg \textbf{Qwen2.5-Coder-0.5B-Instruct} & 0.5B                & \fst{33.9}  & \fst{27.8}  \\
            \midrule
            \multicolumn{4}{c}{\textbf{1B+ Models}}                                                        \\ \midrule
            DS-Coder-1.3B-Instruct                       & 1.3B                & 12.9        & 28.1        \\
            Yi-Coder-1.5B-Chat                           & 1.5B                & 19.9        & 24.9        \\
            \graybg \textbf{Qwen2.5-Coder-1.5B-Instruct} & 1.5B                & \fst{45.4}  & \fst{37.5}  \\
            \midrule
            \multicolumn{4}{c}{\textbf{3B+ Models}}                                                        \\ \midrule
            \graybg \textbf{Qwen2.5-Coder-3B-Instruct}   & 3B                  & \fst{53.2}  & \fst{56.0}  \\
            \midrule
            \multicolumn{4}{c}{\textbf{6B+ Models}}                                                        \\ \midrule
            CodeLlama-7B-Instruct                        & 7B                  & 36.1        & 36.2        \\
            DS-Coder-6.7B-Instruct                       & 6.7B                & 42.6        & 45.1        \\
            CodeQwen1.5-7B-Chat                          & 7B                  & 44.0        & 38.8        \\
            Yi-Coder-9B-Chat                             & 9B                  & 47.5        & 55.6        \\
            DS-Coder-V2-Lite-Instruct                    & 2.4/16B             & 53.0        & 52.9        \\
            \graybg \textbf{Qwen2.5-Coder-7B-Instruct}   & 7B                  & \fst{65.8}  & \fst{65.9}  \\
            \midrule
            \multicolumn{4}{c}{\textbf{13B+ Models}}                                                       \\ \midrule
            CodeLlama-13B-Instruct                       & 13B                 & 47.5        & 41.1        \\
            Starcoder2-15B-Instruct-v0.1                 & 15B                 & 45.5        & 50.9        \\
            \graybg \textbf{Qwen2.5-Coder-14B-Instruct}  & 14B                 & \fst{69.5}  & \fst{79.5}  \\
            \midrule
            \multicolumn{4}{c}{\textbf{20B+ Models}}                                                       \\ \midrule
            CodeLlama-34B-Instruct                       & 34B                 & 48.5        & 47.1        \\
            CodeStral-22B-v0.1                           & 22B                 & 61.3        & 63.5        \\
            DS-Coder-33B-Instruct                        & 33B                 & 47.3        & 50.6        \\
            CodeLlama-70B-Instruct                       & 70B                 & 56.5        & 57.8        \\
            DS-Coder-V2-Instruct                         & 21/236B             & 70.0        & 75.1        \\
            \graybg \textbf{Qwen2.5-Coder-32B-Instruct}  & 32B                 & \fst{75.2}  & \fst{83.4}  \\
            \midrule
            \multicolumn{4}{c}{\grayt \textbf{Closed-APIs}}                                                       \\ \midrule
            \grayt Claude-3.5-Sonnet-20240620            & \grayt -            & \grayt 75.5 & \grayt 81.8 \\
            \grayt Claude-3.5-Sonnet-20241022            & \grayt -            & \grayt 84.4 & \grayt 87.2 \\
            \grayt GPT-4o-mini-2024-07-18                & \grayt -            & \grayt 67.5 & \grayt 78.4 \\
            \grayt GPT-4o-2024-08-06                     & \grayt -            & \grayt 78.6 & \grayt 89.2 \\
            \grayt o1-mini                               & \grayt -            & \grayt 91.6 & \grayt 96.2 \\
            \grayt o1-preview                            & \grayt -            & \grayt 86.5 & \grayt 81.4 \\
            \bottomrule
        \end{tabular}
    }
    \caption{The CRUXEval performance of different instruct models, with \textit{Input-CoT} and \textit{Output-CoT} settings.}
    \vspace{-20pt}
    \label{tab:instruct-cruxeval}
\end{table}

\paragraph{MdEval} Qwen2.5-Coder is further evaluated on the comprehensive multilingual code debugging benchmark MdEval~\citep{mdeval} across 18 languages. Compared to the multilingual code generation benchmark McEval \citep{chai2024mceval}, MdEval provides the buggy code with example test cases (1.2K samples) to LLM for generating the correct code. Figure \ref{fig:instruct-mdeval} demonstrates that the Qwen2.5-Coder-32B-Instruct achieves a comparable or better performance even compared to LLMs with larger model sizes.

\begin{figure}[htbp]
    \centering
    \includegraphics[width=0.9\columnwidth]{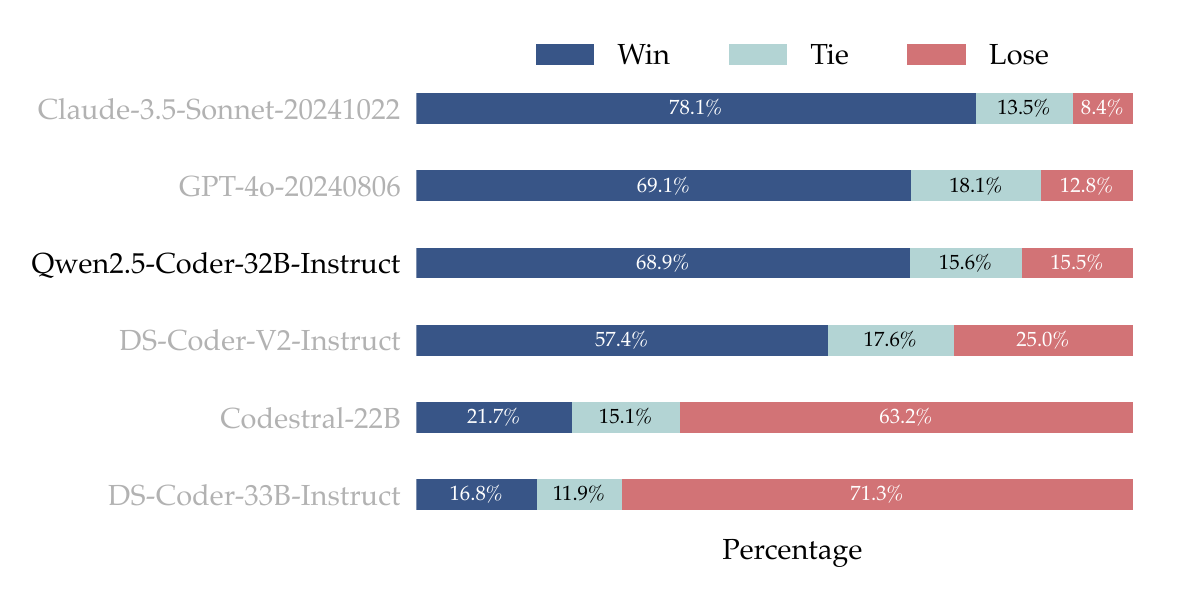}
    \caption{The CodeArena Performance of Qwen2.5-Coder-32B-Instruct compared with popular open-source large code models with similar size.}
    \label{fig:codearena}
\end{figure}

\paragraph{Human Preference Alignment} To evaluate the alignment performance of Qwen2.5-Coder-32B-Instruct with the human preferences, we adopted an internal annotated evaluation benchmark called CodeArena, including nearly 400 human-curated samples. Similar to Chatbot Arena~\citep{chatbot_arena}, we use CodeArena to emulate user code-related prompts in realistic environments. We use GPT-4o as the evaluation model for preference alignment, employing an ``A vs. B win'' evaluation method, which measures the percentage of instances in the test set where the score of A exceeds the score of B. The results in Figure \ref{fig:codearena} demonstrate the advantage of Qwen2.5-Coder-32B-Instruct in preference alignment.

\subsection{Code Reasoning}

\begin{figure}[htbp]
    \centering
    \includegraphics[width=0.8\columnwidth]{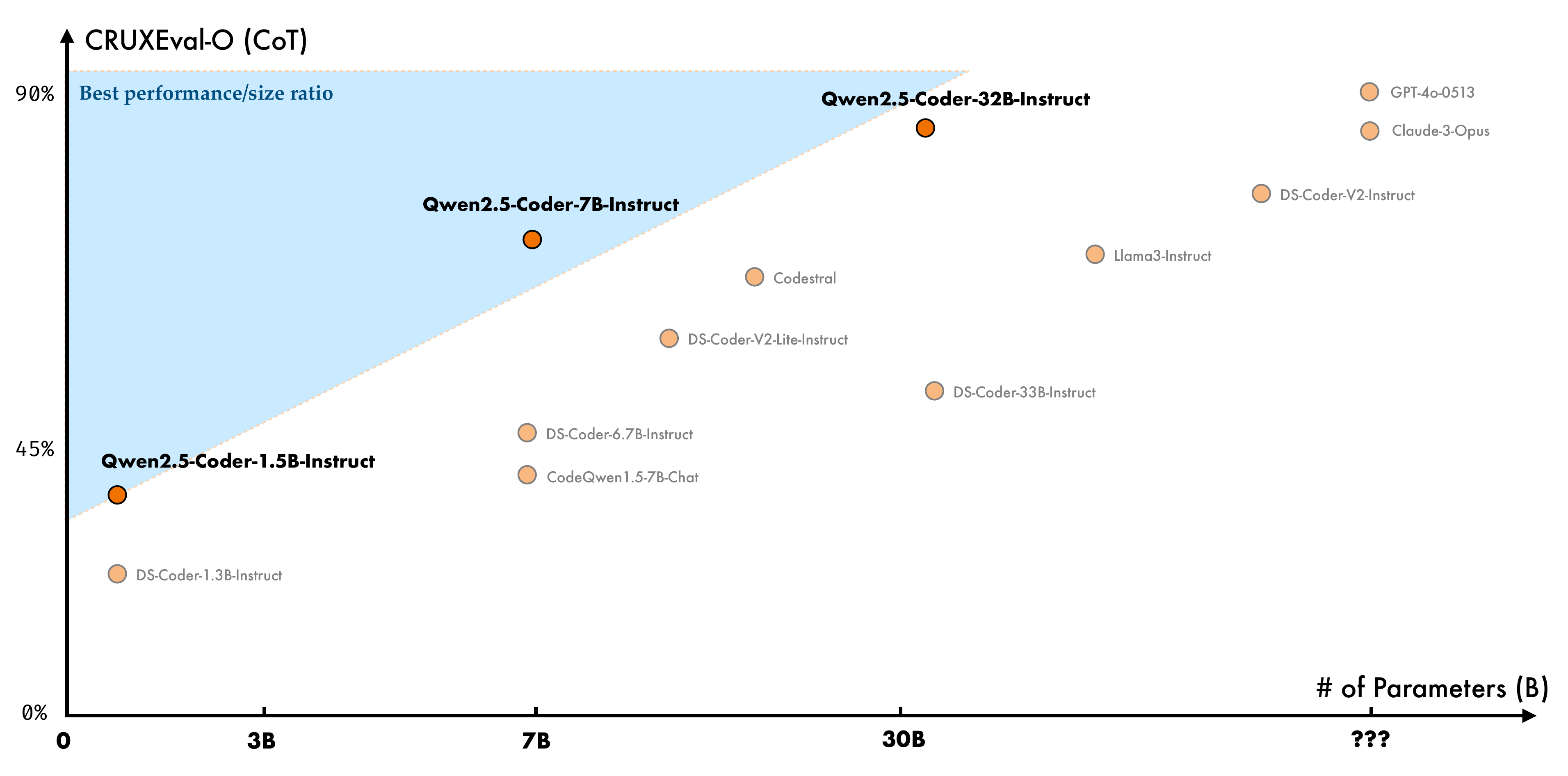}
    \caption{The relationship between model sizes and code reasoning capabilities. The x-axis represents the parameter sizes of different models, and the y-axis indicates the CRUXEval-O (CoT) scores respectively.}
    \label{fig:crux}
\end{figure}

To evaluate the code reasoning capabilities of the Qwen2.5-Coder series instruct models, we conducted an assessment on the CRUXEval \citep{gu2024cruxeval} dataset. As shown in Table \ref{tab:instruct-cruxeval}, the Qwen2.5-Coder-7B-Instruct model achieved Input-CoT and Output-CoT accuracies of 65.8\% and 65.9\%, respectively—demonstrating a substantial improvement over the DS-Coder-V2-Lite-Instruct model, with gains of 12.8\% in Input-CoT accuracy and 13.0\% in Output-CoT accuracy. Additionally, the Qwen2.5-Coder-7B-Instruct model outperformed larger models, including CodeStral-22B and DS-Coder-33B-Instruct, highlighting its advanced code reasoning capabilities despite its smaller size. Notably, our Qwen2.5-Coder-32B-Instruct model achieved accuracies of 75.2\% and 83.4\% on Input-CoT and Output-CoT, respectively, significantly outperforming other open-source code models (including DS-Coder-V2-Instruct) and underscoring its robust performance in code reasoning.

Figure \ref{fig:crux} illustrates the relationship between model sizes and code reasoning capabilities. The Qwen2.5-Coder instruct models stand out for delivering superior code reasoning performance with the fewest parameters, surpassing the results of other open-source large language models by a significant margin.

\begin{table}[h]
    \centering
    \resizebox{0.7\textwidth}{!}{
    \begin{tabular}{lr|cc}
        \toprule
        \multirow{2}{*}{\textbf{Model}}              &
        \multirow{2}{*}{\textbf{Size}}               &
        \multicolumn{2}{c}{\textbf{Aider}}                                                                     \\
                                                     &                 &
        \textit{Pass@1}                              & \textit{Pass@2}                                         \\
        \midrule
        \multicolumn{4}{c}{\textbf{0.5B+ Models}}                                                              \\ \midrule
        \graybg \textbf{Qwen2.5-Coder-0.5B-Instruct} & 0.5B            & \fst{14.3}        & \fst{14.3}        \\
        \midrule
        \multicolumn{4}{c}{\textbf{1B+ Models}}                                                                \\ \midrule
        DS-Coder-1.3B-Instruct                       & 1.3B            & 18.0              & 18.8              \\
        Yi-Coder-1.5B-Chat                           & 1.5B            & 17.3              & 17.3              \\
        \graybg \textbf{Qwen2.5-Coder-1.5B-Instruct} & 1.5B            & \fst{28.6}        & \fst{31.6}        \\
        \midrule
        \multicolumn{4}{c}{\textbf{3B+ Models}}                                                                \\ \midrule
        \graybg \textbf{Qwen2.5-Coder-3B-Instruct}   & 3B              & \fst{33.8}        & \fst{39.1}        \\
        \midrule
        \multicolumn{4}{c}{\textbf{6B+ Models}}                                                                \\ \midrule
        CodeLlama-7B-Instruct                        & 7B              & 1.5               & 1.5               \\
        DS-Coder-6.7B-Instruct                       & 6.7B            & 37.6              & 44.4              \\
        CodeQwen1.5-7B-Chat                          & 7B              & 24.8              & 38.3              \\
        Yi-Coder-9B-Chat                             & 9B              & 45.9              & 54.1              \\
        DS-Coder-V2-Lite-Instruct                    & 2.4/16B         & 44.4              & 52.6              \\
        \graybg \textbf{Qwen2.5-Coder-7B-Instruct}   & 7B              & \fst{55.6}        & \fst{68.4}        \\
        \midrule
        \multicolumn{4}{c}{\textbf{13B+ Models}}                                                               \\ \midrule
        CodeLlama-13B-Instruct                       & 13B             & 1.5               & 1.5               \\
        \graybg \textbf{Qwen2.5-Coder-14B-Instruct}  & 14B             & \fst{58.6}        & \fst{69.2}        \\
        \midrule
        \multicolumn{4}{c}{\textbf{20B+ Models}}                                                               \\ \midrule
        CodeLlama-34B-Instruct                       & 34B             &  1.5             & 1.5              \\
        CodeStral-22B-v0.1                           & 22B             & 36.8              & 51.1              \\
        DS-Coder-33B-Instruct                        & 33B             & 50.4              & 54.5              \\
        CodeLlama-70B-Instruct                       & 70B             & 12.8              & 15.0               \\
        DS-Coder-V2-Instruct                         & 21/236B         & 51.9              & \fst{73.7}         \\
        \graybg \textbf{Qwen2.5-Coder-32B-Instruct}  & 32B             & \fst{60.9} & \fst{73.7}         \\
        \midrule
        \multicolumn{4}{c}{\grayt \textbf{Closed-APIs}}                                                               \\ \midrule
        \grayt Claude-3.5-Sonnet-20240620            & \grayt -        & \grayt 59.4       & \grayt 66.2       \\
        \grayt Claude-3.5-Sonnet-20241022            & \grayt -        & \grayt \fst{71.4} & \grayt 86.5       \\
        \grayt GPT-4o-mini-2024-07-18                & \grayt -        & \grayt 43.6       & \grayt 55.6       \\
        \grayt GPT-4o-2024-08-06                     & \grayt -        & \grayt 56.8       & \grayt 74.4       \\
        \grayt o1-mini                               & \grayt -        & \grayt 49.6       & \grayt 70.7       \\
        \grayt o1-preview                            & \grayt -        & \grayt 69.9       & \grayt \fst{88.0} \\
        \bottomrule
    \end{tabular}
    }
    \caption{
        The code editing ability of different instruct models evaluated by Aider benchmark. The \emph{whole} edit-format was consistently applied across all our experiments.
    }
    \vspace{-10pt}
    \label{tab:instruct-fixing}
\end{table}

\begin{figure}[htbp]
    \centering
    \includegraphics[width=0.75\columnwidth]{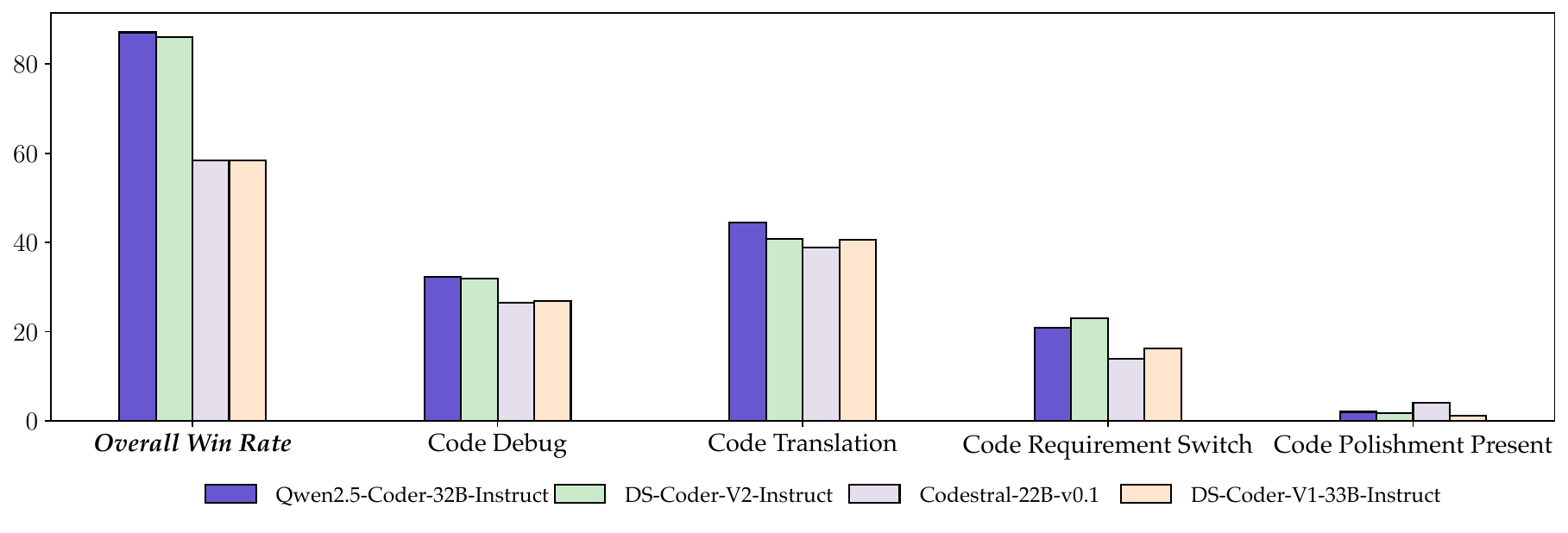}
    \caption{The evaluation results on CodeEditBench.}
    \label{tab:codeeditor}
    \vspace{-12pt}
\end{figure}

\begin{figure}[h!]
    \centering
    \includegraphics[width=1.0\columnwidth]{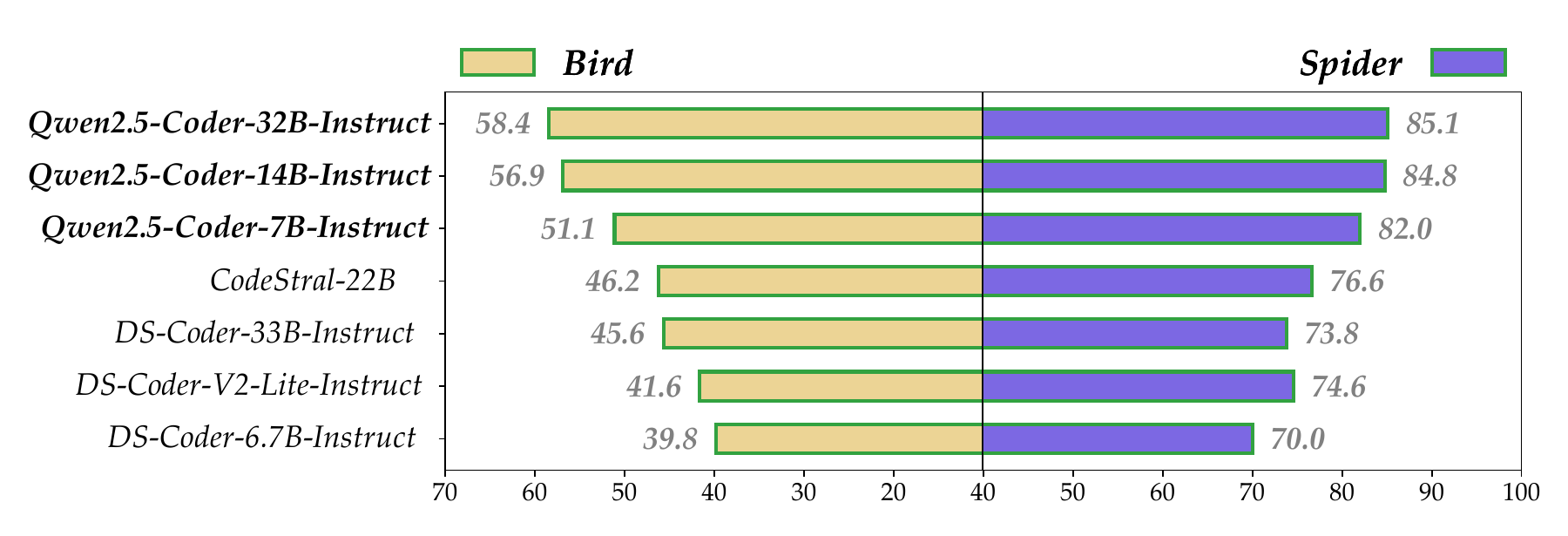}
    \vspace{-1mm}
    \caption{The text-to-SQL evaluation on various instruct code models.}
    \label{fig:instruct-sql}
    \vspace{-10pt}
\end{figure}

\subsection{Code Editing}

\paragraph{Aider} Aider\footnote{\url{https://github.com/paul-gauthier/aider}} has created a code editing benchmark designed to quantitatively measure its collaboration with large language models (LLMs). Drawing from a set of 133 Python exercises sourced from Exercism\footnote{\url{https://github.com/exercism/python}}, the benchmark tests the ability of Aider and LLMs to interpret natural language programming requests and translate them into executable code that successfully passes unit tests. This assessment goes beyond evaluating raw coding proficiency; it also examines how effectively LLMs can edit existing code and format those modifications for seamless integration with Aider’s system, ensuring that local source files can be updated without issues. The comprehensive nature of this benchmark reflects both the technical aptitude of the LLMs and their consistency in task completion. Table \ref{tab:instruct-fixing} highlights the performance of several language models in the Code Editing task.
Among these models, Qwen2.5-Coder-7B-Instruct exhibits exceptional code repair capabilities. Despite its relatively modest scale of 7 billion parameters, it achieves an impressive PASS@1 accuracy of 51.9\%, significantly outperforming comparable models. Remarkably, it also surpasses larger models such as CodeStral-22B and DS-Coder-33B-Instruct , highlighting its remarkable efficiency and effectiveness in code editing tasks. Our Qwen2.5-Coder-32B-Instruct model achieves even higher accuracy, with Pass@1 and Pass@2 rates reaching 60.9\% and 73.7\%, respectively.

\paragraph{CodeEditorBench} 
An effective code assistant must excel in generating code based on given specifications, as well as in modifying or debugging existing code to meet evolving requirements or resolve issues. In evaluating Qwen2.5-Coders proficiency in code modification tasks, we focused on the CodeEditorBench~\citep{codeeditorbench} suite, which assesses performance across four key dimensions: Debugging, Translation, Switching, and Polishing. We employed the same evaluation approach used in the original paper, relying on win rate as the metric for overall performance across diverse problem types. The win rate was computed for each problem category and then averaged across all categories to obtain the overall score. The results in Figure~\ref{tab:codeeditor} show that Qwen2.5-Coder-32B-Instruct achieves a win rate comparable to DS-Coder-V2-Instruct (86.2\% win rate), which features a significantly larger 236 billion parameter scale.

\subsection{Text-to-SQL}
SQL is one of the essential tools in daily software development and production, but its steep learning curve often hinders free interaction between non-programming experts and databases. To address this issue, the Text-to-SQL task was introduced, aiming for models to automatically map natural language questions to structured SQL queries. Previous improvements in Text-to-SQL focused primarily on structure-aware learning, domain-specific pre-training, and sophisticated prompt designs.


Thanks to the use of finely crafted synthetic data during both pre-training and fine-tuning, we significantly enhanced \qwencoder's capability in Text-to-SQL tasks. We selected two well-known benchmarks, Spider \citep{yu2019spiderlargescalehumanlabeleddataset} and BIRD \citep{li2023llmservedatabaseinterface}, for comprehensive evaluation.
To ensure a fair comparison between \qwencoder and other open-source language models on this task, we used a unified prompt template as input, following the work of  \citet{chang2023prompt}. The evaluation prompt consists of table representations aligned with database instructions, examples of table content, optional additional knowledge, and natural language questions. This standardized prompt template minimizes biases that may arise from prompt variations. As shown in Figure~\ref{fig:instruct-sql}, \qwencoder outperforms other code models of the same size on the Text-to-SQL task.



\begin{figure}[h!]
    \centering
    \includegraphics[width=0.75\columnwidth]{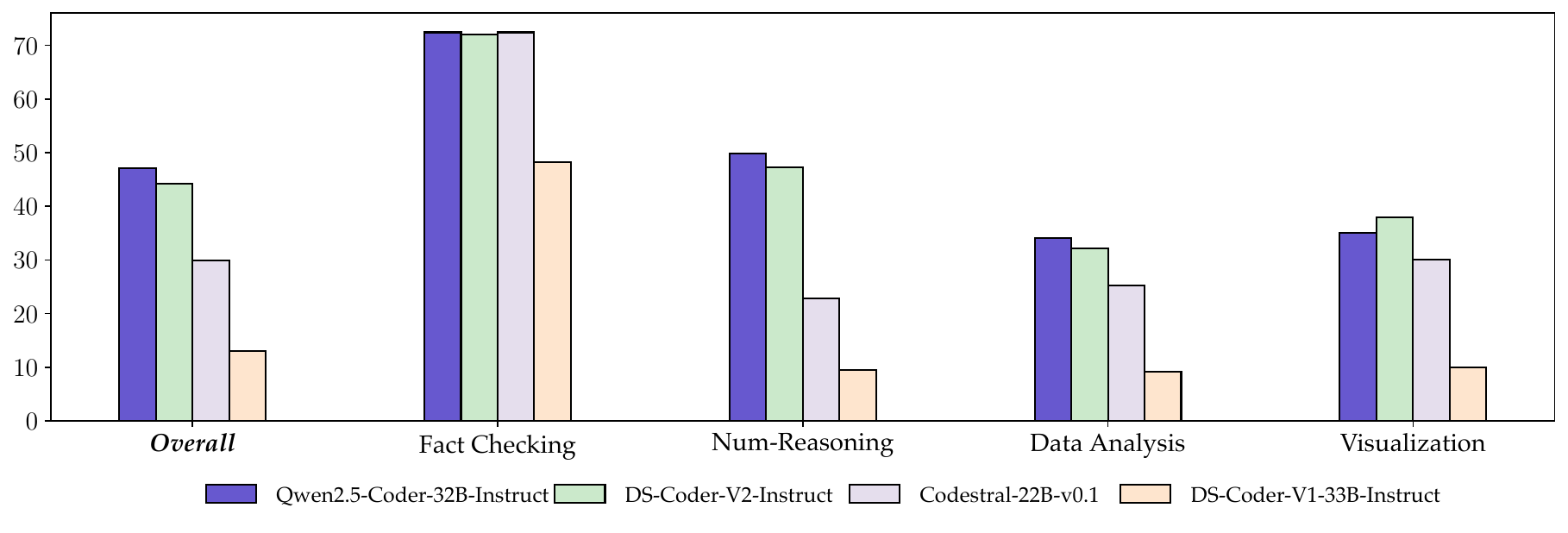}
    \vspace{-1mm}
    \caption{The table understanding evaluation on TableBench.}
    \vspace{-10pt}
    \label{fig:instruct-table}
\end{figure}


\begin{table}
 \centering
 \resizebox{\textwidth}{!}{
 \begin{tabular}{lr|ccccccc}
 \toprule
 \textbf{Model} & \textbf{Size} &
 \textbf{MATH} & \textbf{GSM8K} & \textbf{GaoKao2023en} & \textbf{OlympiadBench} & \textbf{CollegeMath} & \textbf{AIME24} \\
 \midrule
 DS-Coder-V2-Lite-Instruct & 2.4/16B & 61.0 & 87.6 & 56.1 & 26.4 & 39.8 & 6.7 \\
 DS-Coder-V2-Instruct & 21/236B & 74.2 & \fst{94.5} & 65.7 & 37.8 & 45.9 & 6.7 \\
 \midrule
 \graybg \textbf{Qwen2.5-Coder-3B-Instruct} & 3B & 58.1 & 80.7 & 48.8 & 23.6 & 39.7 & 6.7 \\
 \graybg \textbf{Qwen2.5-Coder-7B-Instruct} & 7B & 66.8 & 86.7 & 60.5 & 29.8 & 43.5 & 10.0 \\
 \graybg \textbf{Qwen2.5-Coder-14B-Instruct} & 14B & 66.8 & 94.2 & 66.0 & 40.1 & 47.3 & 10.0 \\
 \graybg \textbf{Qwen2.5-Coder-32B-Instruct} & 32B & \fst{76.4} & 93.0 & \fst{68.3} & \fst{42.5} & \fst{47.7} & \fst{20.0} \\
 \bottomrule
 \toprule
 \textbf{Model} & \textbf{Size} &
 \textbf{AMC23} & \textbf{MMLU} & \textbf{MMLU-Pro} & \textbf{IFEval} & \textbf{CEval} & \textbf{GPQA} \\
 \midrule
 DS-Coder-V2-Lite-Instruct & 2.4/16B & 40.4 & 42.5 & 60.6 & 38.6 & 60.1 & 27.6 \\
 DS-Coder-V2-Instruct & 21/236B & 52.5 & 76.7 & \fst{65.6} & 40.9 & \fst{73.4} & \fst{44.3} \\
 \midrule
 \graybg \textbf{Qwen2.5-Coder-3B-Instruct} & 3B & 25.0 & 56.5 & 35.2 & 44.2 & 53.9 & 28.3 \\
 \graybg \textbf{Qwen2.5-Coder-7B-Instruct} & 7B & 42.5 & 68.7 & 45.6 & 58.6 & 61.4 & 35.6 \\
 \graybg \textbf{Qwen2.5-Coder-14B-Instruct} & 14B & 50.0 & 71.7 & 55.6 & 66.5 & 66.2 & 36.8 \\
 \graybg \textbf{Qwen2.5-Coder-32B-Instruct} & 32B & \fst{55.0} & \fst{77.6} & 62.3 & \fst{79.9} & 68.9 & 41.8 \\
 \bottomrule
 \end{tabular}
 }
 \caption{The performance of math and general.}
 \label{tab:instruct-math-general}
 \vspace{-10pt}
\end{table}

\subsection{Math Reasoning and General Natural Language}

In this section, we provide a comparative analysis of the performance between our Qwen2.5-Coder series models and the DS-Coder-V2 series models, with a focus on both mathematical computation and general natural language processing tasks. The results in Table \ref{tab:instruct-math-general} highlight the versatility of the Qwen2.5-Coder series, which excels not only in complex coding tasks but also in advanced general-purpose tasks, setting it apart from its competitors.

\subsection{Table Understanding}
To evaluate the understanding capabilities of structured data, we further evaluate the Qwen2.5-Coder on a comprehensive and complex benchmark TableBench~\citep{tablebench}, which includes 18 fields within four major categories of table question answering (TableQA) capabilities. We compare Qwen2.5-Coder with other LLMs under the textual chain-of-thought (TCoT) setting. Figure \ref{fig:instruct-table} demonstrates that Qwen2.5-Coder-32B-Instruct gets the best performance 45.1 on TableBench.

\section{Discussion: Scaling is All You Need}
\begin{figure}[h!]
    \centering
    \includegraphics[width=1.0\columnwidth]{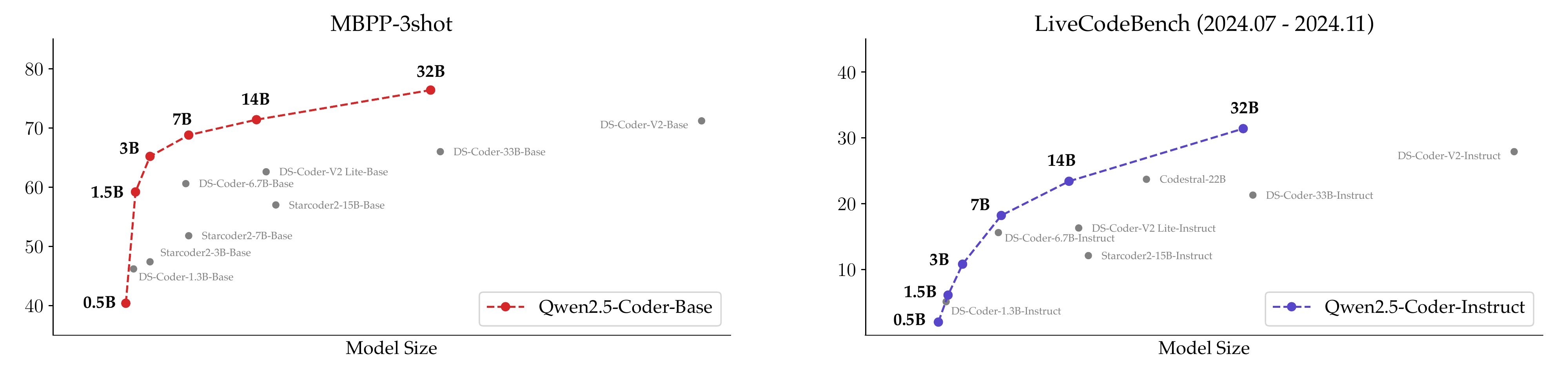}
    \caption{The evaluation results of Qwen2.5-Coder models with different sizes on MBPP-3shot and LiveCodeBench.}
    \vspace{-10pt}
    \label{fig:model_scaling}
\end{figure}

In Figure \ref{fig:model_scaling}, We present a comparison of different sizes of Qwen2.5-Coder with other open-source LLMs on MBPP-3shot and LiveCodeBench. For the base LLM, we choose MBPP-3shot as the evaluation metric. Our extensive experiments show that MBPP-3shot is more suitable for evaluating base models and correlates well with the actual performance of the models. For the instruction model, we select the latest 4 months of LiveCodeBench (2024.07$\sim$2024.11) questions as the evaluation to strictly avoid test data contamination, truly reflecting the OOD capabilities of the LLM. There is a positive correlation between model size and model performance, and Qwen2.5-Coder has achieved state-of-the-art performance across all sizes, encouraging us to continue exploring larger sizes of code LLM.

\section{Conclusion}

This work introduces Qwen2.5-Coder, the latest addition to the Qwen series. Built upon Qwen2.5, a top-tier open-source LLM, Qwen2.5-Coder has been developed through extensive pre-training and post-training of Qwen2.5-0.5B/1.5B/3B/7B/14B/32B on large-scale datasets. To ensure the quality of the pre-training data, we have curated a dataset by collecting public code data and extracting high-quality code-related content from web texts, while filtering out low-quality data using advanced classifiers. Additionally, we have constructed a meticulously designed instruction-tuning dataset to transform the base code LLM into a strong coding assistant.

Looking ahead, our research will focus on exploring the impact of scaling up code LLMs in terms of both data size and model size. We will also continue to enhance the reasoning capabilities of these models, aiming to push the boundaries of what code LLMs can achieve.

\clearpage

\bibliography{colm2024_conference}
\bibliographystyle{colm2024_conference}


\end{document}